\pdfoutput=1
\documentclass{article}
\usepackage[preprint]{colm2026_conference}

\usepackage{etoolbox}
\usepackage{microtype}
\usepackage{hyperref}
\usepackage{url}
\usepackage{booktabs}
\usepackage{graphicx}
\usepackage{lineno}

\definecolor{darkblue}{rgb}{0, 0, 0.5}
\hypersetup{colorlinks=true, citecolor=darkblue, linkcolor=darkblue, urlcolor=darkblue}

\title{What Is the Minimum Architecture for Prolepsis? Early Irrevocable Commitment Across Tasks in Small Transformers}

\author{\'{E}ric Jacopin\\
Cosmic AI\\
\texttt{eric.jacopin@protonmail.com}}

\begin{document}

\ifcolmsubmission
\linenumbers
\fi

\maketitle

\begin{abstract}
When do transformers commit to a decision, and what prevents them from correcting it? We introduce \textbf{prolepsis}: a transformer commits early, task-specific attention heads sustain the commitment, and no layer corrects it. Replicating \citeauthor{lindsey2025biology}'s (\citeyear{lindsey2025biology}) planning-site finding on open models (Gemma~2 2B, Llama~3.2 1B), we ask five questions. (Q1)~Planning is invisible to six residual-stream methods; among those tested, only CLT-based steering succeeds. (Q2)~The single-site spike replicates in shape, at the final prompt token (Anthropic's site is the newline; see the Note added below). (Q3)~Specific attention heads route the decision to the output, filling a gap flagged as invisible to attribution graphs. (Q4)~The evidence is consistent with search within ${\leq}16$ layers and commitment beyond, a two-model hypothesis. (Q5)~Factual recall shows the same motif at a different network depth, with zero overlap between recurring planning heads and the factual top-10. Prolepsis recurs across tasks in the decoder-only models tested: the template is shared, the routing substrates differ. All experiments run on a single consumer GPU (16\,GB VRAM).
\end{abstract}

\paragraph{Note added (v2, July 2026).} This version corrects and re-scopes v1 in six ways, following the COLM 2026 reviews and subsequent experiments with this implementation (to be reported separately). (i)~Terminology: what v1 calls ``the planning site'' is the \emph{final prompt token}; Anthropic's planning site is the newline between lines. Subsequent position sweeps, a feature census at the newline, and a full-line composition variant show the newline is causally inert in these models: v1's replication claim holds for position specificity, not for site identity, and we now say \emph{commitment site}. (ii)~An exploratory detection analysis in the reference implementation read the CLT encoder from the post-MLP residual; under the correct pre-MLP hook (settled by MLP-output reconstruction) its reported activations are artifacts. No causal result in this paper depends on it. (iii)~Three citations are corrected (Meng et al.; Vig et al.; Moore et al.). (iv)~Main-text and appendix numbers are relabeled where v1 mixed per-pair and best-of-sweep values. (v)~Claims are re-scoped: ``first'' is dropped \citep{hanna2026latent}, ``CLTs are necessary'' is limited to the six methods tested \citep{maar2026plan}, the depth threshold is downgraded to a hypothesis, and ``architectural'' is scoped to the decoder-only models tested. (vi)~The term \emph{prolepsis} is now defined against its established English senses at first use.

%% ============================================================
\section{Introduction}
\label{sec:intro}

\citet{lindsey2025biology} reported a striking finding: when Claude 3.5 Haiku writes rhyming poetry, cross-layer transcoder (CLT) features for candidate rhyme words pre-activate at the newline token between lines, a ``planning site'' where the model decides the rhyme ending before generating the words that lead to it. Suppress\,+\,inject interventions on these features redirect the line ending, but only when applied at the planning site; the same intervention at other token positions has no effect (Figure~13 of ``On the Biology of a Large Language Model''). CLTs are descendants of sparse dictionary learning applied to transformer activations \citep{bricken2023monosemanticity,templeton2024scaling,cunningham2023sparse}: they decompose the residual stream (the running sum of layer outputs that serves as the transformer's main communication channel; \citealp{elhage2021mathematical}) into interpretable, monosemantic features that can be individually suppressed or injected.

This finding was demonstrated on a proprietary model, with proprietary weights, using proprietary interpretability tools. No external researcher can inspect the model or reproduce the experiment on the same system. Until reproduced on independently trained models with independently built infrastructure, it remains a single-lab observation. Meanwhile, mechanistic investigations of planning in neural networks have focused on trained search \citep{lehnert2024beyond}, recurrent planning circuits \citep{taufeeque2024planning,taufeeque2025path}, or behavioral evaluation \citep{moore2025planning}; none has localized a planning circuit inside a pretrained language model using sparse features.

We provide that independent verification, and discover something broader in the process. Our investigation proceeds as a series of questions, each answer raising the next.

\textbf{Q1: Can you observe planning without CLTs?} No; six standard steering methods all fail; among the methods we tested, CLT-based steering is the only lens through which the planning circuit is visible (Section~\ref{sec:invisible}).
\textbf{Q2: Does planning-site localization replicate?} Yes in shape: a single-position spike replicates on Gemma~2 2B and Llama~3.2 1B, at the final prompt token (Section~\ref{sec:replication}; see the Note added).
\textbf{Q3: How does the decision travel to the output?} Through attention routing heads, led by L21:H5, filling a gap flagged as invisible to attribution graphs (Section~\ref{sec:routing}).
\textbf{Q4: What is the minimum depth?} The evidence is consistent with search within ${\leq}16$ layers and commitment beyond, a hypothesis from a two-model comparison (Section~\ref{sec:commitment}).
\textbf{Q5: Is this specific to planning?} No; factual recall shows the same motif at a different network depth, with zero overlap between recurring planning heads and the factual top-10 (Section~\ref{sec:prolepsis}).

We term this motif \textbf{prolepsis} (Greek: ``a taking beforehand''). Prolepsis is an English word with several established senses: in rhetoric, answering an objection before it is raised; in narratology, presenting a future event as already accomplished (the flash-forward); in Epicurean philosophy, a preconception that guides recognition. Common to all is the anticipatory fixing of something that does not yet exist. We use this shared core in its goal-directed form: the model treats the outcome as already decided and produces the intervening computation under that commitment. Technically, prolepsis here names the structural motif in which a transformer commits to a decision at early layers, and attention routing heads sustain that commitment to the output without correction, with three properties: early commitment, sustained propagation through routing heads (at task-dependent network depths), and irrevocability: no layer reverses the committed answer, even when it is wrong.

\paragraph{Contributions.} We make six contributions: (1)~an independent replication of single-site localization \citep{lindsey2025biology} on open models with open CLTs at three scales, at the final prompt token rather than Anthropic's newline (Note added); (2)~a demonstration that planning is invisible to the six residual stream methods we tested, establishing CLT-based steering as necessary among them; (3)~identification of the attention routing heads (L21:H5 family) that carry planning decisions, filling the gap flagged as invisible to attribution graphs; (4)~a minimum-depth hypothesis for commitment (search ${\leq}16$ layers, commitment ${>}16$), consistent with causal evidence from layer suppression on a two-model comparison; (5)~discovery of prolepsis as a cross-task architectural motif, with planning routing through mid-layer heads and factual routing through late-layer heads (zero overlap at the aggregate level); and (6)~full reproducibility on a single consumer GPU (16\,GB VRAM).

\paragraph{Organization.} Section~\ref{sec:related} positions our work relative to prior results. Section~\ref{sec:setup} describes the models, CLTs, tasks, and feature discovery protocol. Sections~\ref{sec:invisible}--\ref{sec:prolepsis} are structured as five research questions (Q1--Q5), each answered by the data presented in that section, and each answer raising the next question. Section~\ref{sec:discussion} discusses implications, limitations, and the CLT coverage constraint.

%% ============================================================
\section{Related work}
\label{sec:related}

\paragraph{Planning in neural networks.} \citet{taufeeque2024planning} identified ``path channels'' in a recurrent network trained to play Sokoban; \citet{taufeeque2025path} extended this to plan extension kernels. Both study networks \emph{trained for planning}. \citet{lehnert2024beyond} trained transformers on A* search traces. \citet{moore2025planning} evaluated planning theory of mind behaviorally. None localizes a planning circuit inside a pretrained language model. \citet{lindsey2025biology} were the first to do so on a proprietary model; we provide an independent replication on open models. Concurrently, \citet{hanna2026latent} reproduce rhyme planning on Qwen3 models with CLTs and argue that latent planning emerges with scale, and \citet{maar2026plan} shift rhyming behavior with contrastive steering vectors at the behavioral level.

\paragraph{Factual recall and knowledge localization.} Factual associations are localized in MLP layers \citep{meng2022locating,dai2022knowledge} and follow a pipeline of attribute extraction and subject enrichment \citep{geva2023dissecting}. We extend this by showing that factual recall and planning use fundamentally different computational substrates within the same model: residual stream directions vs.\ sparse features and attention routing.

\paragraph{Sparse features, steering, and circuits.} Sparse autoencoders \citep{bricken2023monosemanticity,cunningham2023sparse} decompose activations into interpretable features, scaled to large models by \citet{templeton2024scaling}. \citet{lindsey2025biology} used cross-layer transcoders (CLTs) to discover the planning-site phenomenon we replicate. \citet{marks2024sparse} introduced sparse feature circuits for causal graph tracing. Separately, activation addition \citep{turner2023activation}, representation engineering \citep{zou2023representation}, causal mediation \citep{vig2020investigating}, causal abstraction \citep{geiger2021causal}, and automated circuit discovery \citep{conmy2023automated} all operate on the residual stream. Our negative result, that all residual stream methods fail for planning while succeeding for factual recall, identifies a boundary of this paradigm.

%% ============================================================
\section{Experimental setup}
\label{sec:setup}

\paragraph{Models and CLTs.} We test every model for which open CLTs currently exist and that fits on consumer hardware. All three CLTs were released by mntss on HuggingFace\footnote{\url{https://huggingface.co/mntss}} and are, at the time of writing, the only publicly available CLTs for any open-weights language model. This was the set available when these experiments were run; CLTs for other models (e.g., BlueLightAI's for Qwen3) have since appeared.

\smallskip
\noindent
\small
\begin{center}
\begin{tabular}{@{}ll@{}}
\toprule
\textbf{Component} & \textbf{Configuration} \\
\midrule
Gemma 2 2B & 26 layers, 2304 hidden, BF16, base (Google DeepMind) \\
Llama 3.2 1B & 16 layers, 2048 hidden, BF16, base (Meta) \\
CLT (Gemma, 426K) & 16{,}384 features/layer $\times$ 26 layers \\
CLT (Gemma, 2.5M) & 98{,}304 features/layer $\times$ 26 layers \\
CLT (Llama, 524K) & 32{,}768 features/layer $\times$ 16 layers \\
Hardware & Single consumer GPU, 16\,GB VRAM \\
Framework & Custom (open-source upon publication) \\
\bottomrule
\end{tabular}
\end{center}
\normalsize
\smallskip

Our CLT implementation was validated against the Python Circuit Tracer reference: 90/90 top-10 features match (max relative error $1.2 \times 10^{-6}$), with KL divergence of 0.070 under active CLT injection.

\paragraph{Rhyme planning task.} Gemma~2 2B produces 0\% rhymes on single-line prompts but 78\% with a priming couplet that establishes a different rhyme ending. We use four-line completion prompts: lines 1--2 set a rhyme ending, line 3 introduces a target ending, and the model completes line~4. Wrapping target words in \texttt{<strong>} HTML tags increases the rhyme rate from 78\% to 81\%; the model associates HTML emphasis with phonologically salient line-ending words, direct evidence of learning from web-published poetry. Llama requires a separate prompt set due to different CLT vocabulary coverage.

\paragraph{Factual recall task.} We use Transluce's activation patching dataset \citep{li2025transluce}, generated on Llama~3.1 8B with CounterFact prompts \citep{meng2022locating}. From 1{,}866 unique prompt pairs, we screen for cases where Llama~3.2 1B correctly predicts both the original and counterfactual answers, yielding 89 gold pairs with maximum causal power.

\paragraph{CLT feature discovery.} We scanned all 425{,}984 features against all 256{,}768 token embeddings via chunked GPU matmul, yielding 287 features with clean English words as top-1 token. Cross-referencing with the CMU Pronouncing Dictionary \citep{weide1998cmu} reveals features cluster by phonetic ending: 35 rhyme groups for Gemma, 12 for Llama. The 2.5M CLT resolves individual words (209 words, mean decoder rank 1.0, zero cross-group contamination). Full protocol in Appendix~\ref{app:discovery}.

%% ============================================================
\section{Q1: Can planning be observed without CLTs?}
\label{sec:invisible}

Before discovering the bottom-up CLT approach, we applied six standard interpretability methods to redirect rhyme endings in Gemma~2 2B: max-activation probes, contrastive steering vectors \citep{turner2023activation,zou2023representation}, cosine-filtered planning-site injection, multi-layer clamping, contrastive word probes, and causal activation patching \citep{vig2020investigating,geiger2021causal,conmy2023automated}. All six fail (0\% target hit; full results in Appendix~\ref{app:steering}).

To confirm the methods work, we applied the same contrastive steering protocol to factual recall. Injecting (ResidPost$_\mathrm{france}$ $-$ ResidPost$_\mathrm{germany}$) at any layer redirects the output from ``Paris'' to ``Berlin'' with 39.3\% probability. The same protocol applied to 20 rhyme groups yields ${<}0.5\%$ for every group.

This asymmetry reveals a structural fact. Factual recall is encoded as a \emph{direction} in the residual stream \citep{elhage2021mathematical} with a hard attractor basin: perturbations below ${\sim}1.2\times$ the contrastive distance are fully absorbed within 1--2 layers (81\% absorption in Llama, 92\% in Gemma). This is consistent with prior work showing factual associations are localized in specific layers and MLP modules \citep{meng2022locating,geva2023dissecting,dai2022knowledge}. Planning has no such direction. It is encoded in sparse CLT feature activations and routed through attention heads, a fundamentally different computational regime. The residual stream does not distinguish prompts with different rhyme-setting words (within-group KL ${<}$ 0.02).

Among the methods we tested, CLTs are not merely a convenient tool for observing planning; they are the only one that succeeds. Without sparse feature decomposition, the planning circuit is invisible to all six residual stream approaches.

\paragraph{A1.} No. Planning is invisible to all six residual stream methods we tested. It requires sparse feature decomposition to observe.

%% ============================================================
\section{Q2: Does planning-site localization replicate on open models?}
\label{sec:replication}

\begin{figure}
\begin{center}
\includegraphics[width=0.75\linewidth]{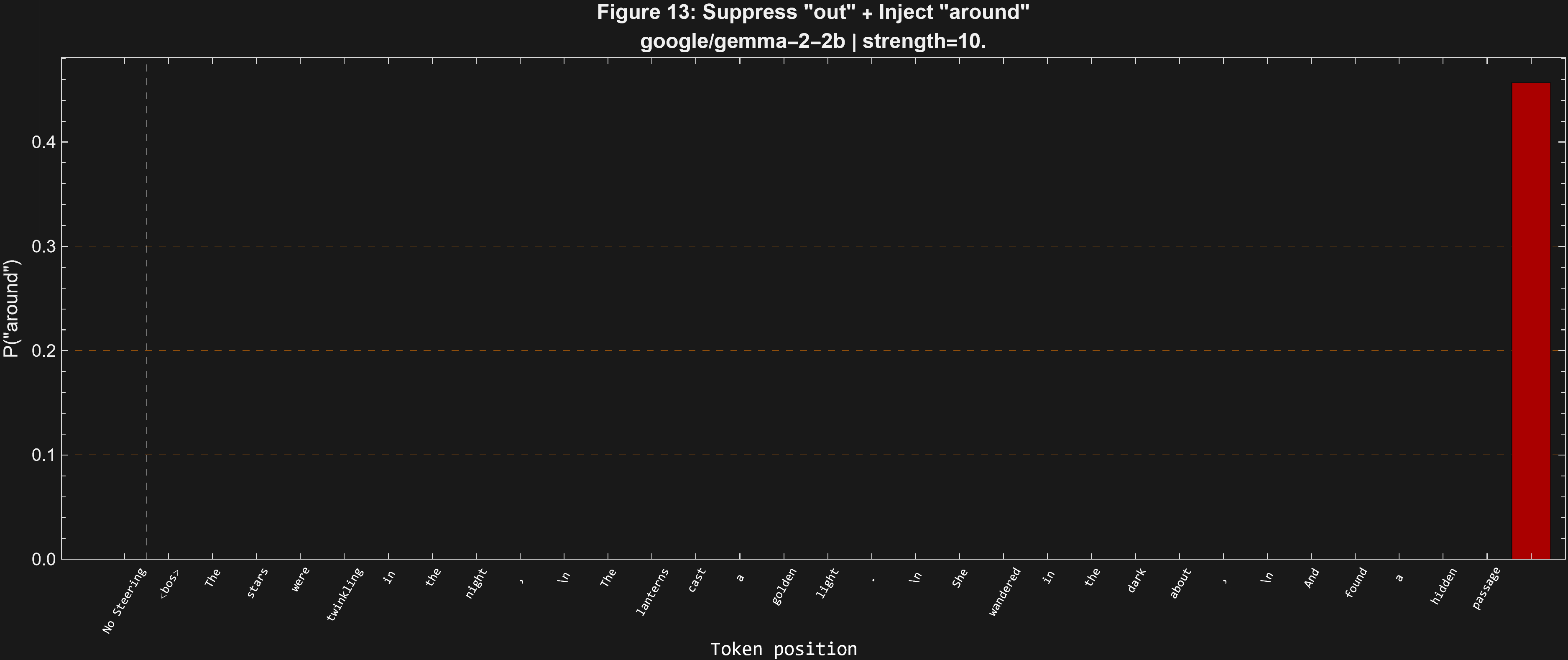}
\end{center}
\caption{Position sweep on Gemma~2 2B (426K CLT): suppress ``out'' features, inject ``around'' at each token position, measure P(``around''). Flat baseline at ${\sim}10^{-8}$ across all positions; sharp spike to 0.457 at the planning site (last position). This replicates the position-specificity of Anthropic's Figure~13 \citep{lindsey2025biology}; the effective site here is the final prompt token, not Anthropic's newline (Note added).}
\label{fig:spike}
\end{figure}

We follow the suppress\,+\,inject position-sweep protocol of \citet{lindsey2025biology}: suppress all CLT features from the natural rhyme group (negative strength, all downstream layers), inject a single feature from an alternative group (positive strength), and sweep the injection position across every token. We measure P(injected word) at each position. A concrete example illustrates the setup. Given the prompt:

\smallskip
\noindent\textit{The stars were twinkling in the night,} \\
\textit{The lanterns cast a golden light.} \\
\textit{She wandered in the dark \textbf{about},} \\
\textit{And found a hidden passage \rule{1cm}{0.4pt}}
\smallskip

\noindent the natural rhyme group is ``-out'' (about, out, shout). We suppress the ``out'' CLT features and inject the ``around'' feature at each token position. At all 31 positions except the last, P(``around'') $\approx 5 \times 10^{-8}$. At the planning site (position 31, the space before generation), it jumps to 0.457 (Figure~\ref{fig:spike}). On the best-performing prompt from the 136-pair sweep, P(``around'') reaches 0.483. The model's rhyme decision is localized to a single token position.

%\smallskip
\paragraph{Gemma 2 2B, 426K CLT (group-level).} 136 pairs (4 prompts $\times$ 34 alternative groups). 70\% of pairs (95/136) show maximum P(inject word) at the planning site. Best redirect: P(``around'') $= 0.483$ at position~31, up from a baseline of $4.5 \times 10^{-8}$, a ten-million-fold increase.

The layer-depth gradient confirms the mechanism: earlier injection leaves more downstream layers to propagate the signal. L16 produces a $160{,}000\times$ ratio, L19 produces $1{,}300\times$, L22 produces $770\times$, L25 produces $610\times$.

\paragraph{2.5M CLT (word-level).} 264 pairs; 62.1\% localization at the commitment site (164/264; Appendix~
ef{app:sweeps}). Best redirect: 52.2\% (``black'' at L25). Best ratio: $3.78 \times 10^{12}$ (``kind'' at L25). Every rhyme word has its own dedicated feature. Runs on 16\,GB VRAM via stream-and-free shard loading.

%\smallskip
\paragraph{Llama 3.2 1B.} 44 pairs (4 prompts $\times$ 11 alternative groups). All strong injections peak at the planning site. Best redirect for the preset worked pair: P(``that'') $= 0.777$ at L14 ($133{,}879\times$ ratio over that pair's baseline); an initial 47-pair sweep, retained in Appendix~\ref{app:sweeps}, reached $0.834$. The spike shape is structurally identical to Gemma's. Llama's vocabulary is sparser: 28 usable words in 12 groups vs.\ Gemma's 98 in 35.

\begin{table}
\small
\begin{center}
\begin{tabular}{lccc}
\toprule
& \textbf{Claude 3.5 Haiku} & \textbf{Gemma 2 2B} & \textbf{Llama 3.2 1B} \\
\midrule
Steering success & 70\% (25 poems) & 52.2\% (2.5M) & 77.7\% \\
Localization & Not reported$^\dagger$ & 62--70\% & Yes (all strong) \\
Sweep shape & Spike at newline & Spike at planning site & Spike at planning site \\
CLT resolution & 30M & 426K / 2.5M & 524K \\
Elicitation & Low & High & High \\
\bottomrule
\end{tabular}
\end{center}
\caption{Cross-model comparison. $^\dagger$\citet{lindsey2025biology} report a 70\% steering success rate (injected word appears in generated line, $n=25$ poems) and note planned-word features in ``about half'' of investigated poems, but do not report a position-sweep localization rate comparable to ours.}
\label{tab:replication}
\end{table}

\paragraph{A2.} Yes. The spike shape replicates on both models with structurally identical geometry (Table~\ref{tab:replication}). Gemma 426K localizes at 70\% (95/136; 95\% CI [62\%, 78\%]). One difference: Anthropic's planning site is the newline token between lines (the model commits many tokens before the rhyme word), while ours is the last token before generation. The spatial gap between commitment and output is shorter in our setup; the localization pattern is the same. Subsequent experiments show this difference is substantive: the same intervention applied at the newline itself is causally inert in these models (Note added). The signal is sparse and hard-won in small models, a finding about the scaling of planning, not a limitation. This raises the next question: how does the decision at the planning site propagate to the output?

%% ============================================================
\section{Q3: How does the planning decision travel to the output?}
\label{sec:routing}

CLT features modify the residual stream at the rhyme-word position, but the model predicts the next word at the last token. Something must connect these positions. Prior work has identified functionally specialized attention heads (induction heads for in-context learning \citep{olsson2022context}, factual recall heads \citep{geva2023dissecting}) but no attention head has been identified as carrying a \emph{planning} decision. We measured attention pattern changes across all 208 heads (26 layers $\times$ 8 heads) in Gemma~2 2B during the same suppress\,+\,inject intervention.

\begin{figure}
\begin{center}
\begin{minipage}{0.19\linewidth}
\centering
\includegraphics[width=\linewidth]{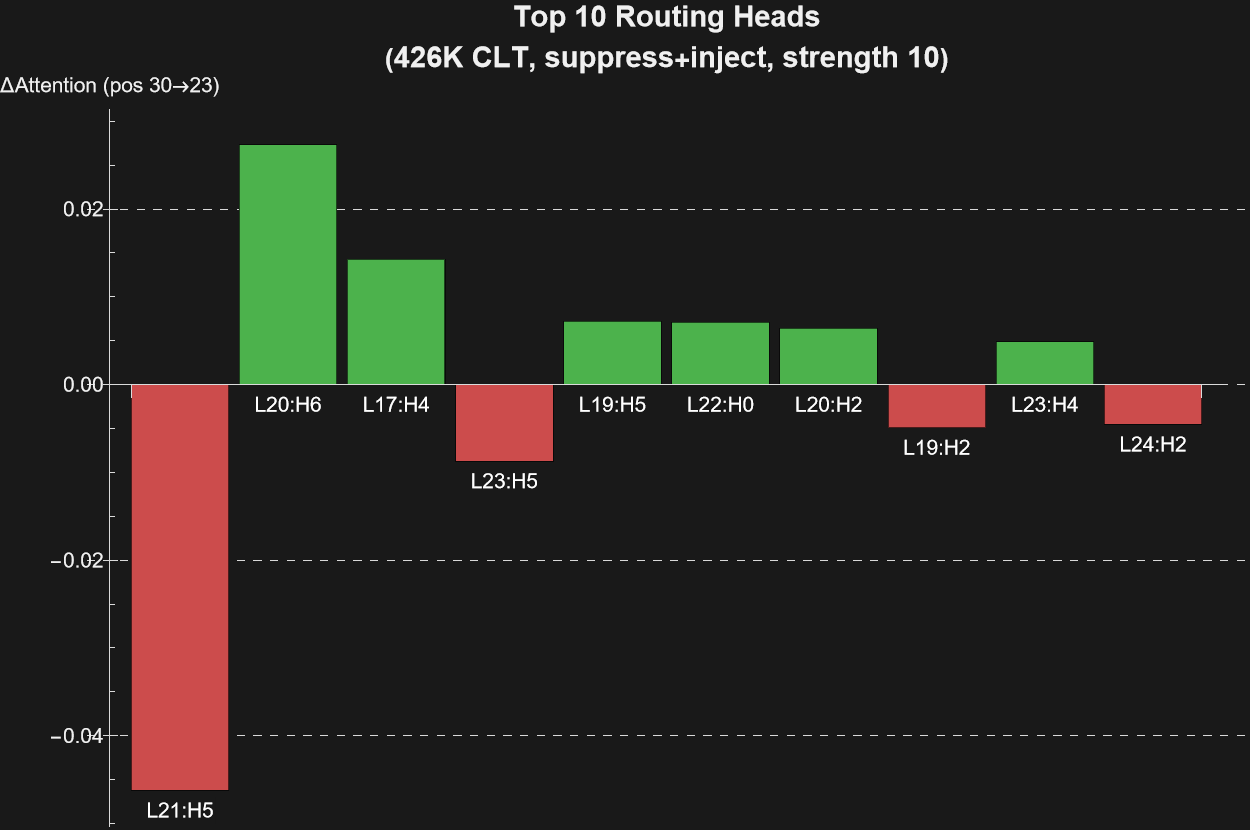}
\smallskip
{\scriptsize (a) Gemma 2 2B}
\end{minipage}
\hfill
\begin{minipage}{0.19\linewidth}
\centering
\includegraphics[width=\linewidth]{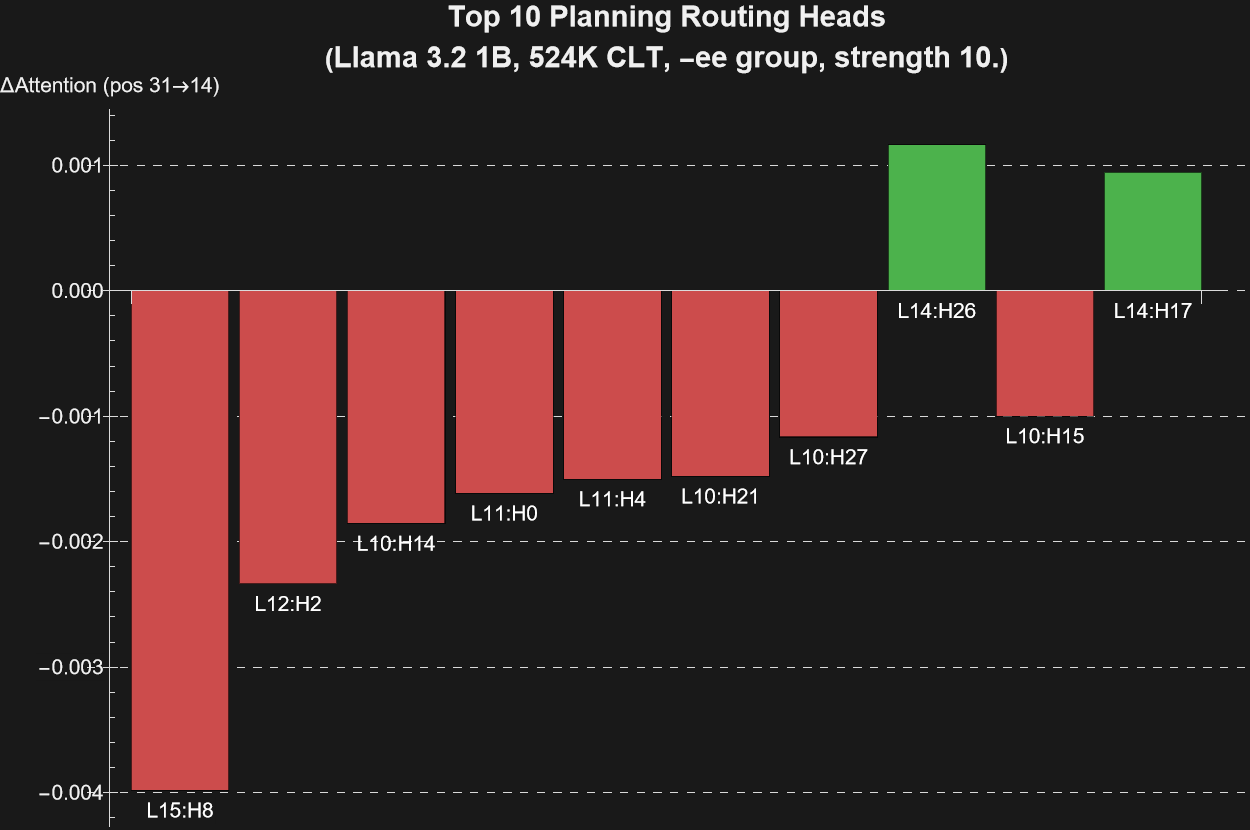}
\smallskip
{\scriptsize (b) Llama -ee}
\end{minipage}
\hfill
\begin{minipage}{0.19\linewidth}
\centering
\includegraphics[width=\linewidth]{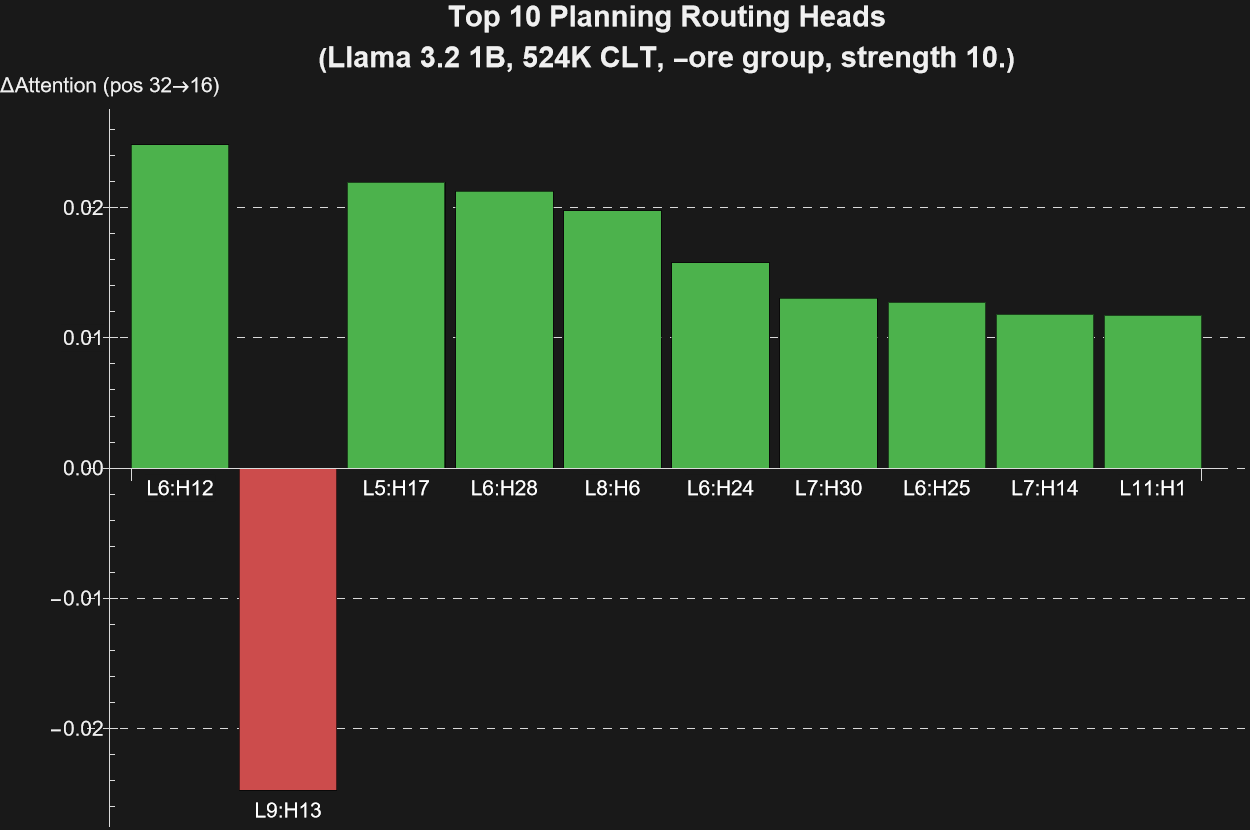}
\smallskip
{\scriptsize (c) Llama -ore}
\end{minipage}
\hfill
\begin{minipage}{0.19\linewidth}
\centering
\includegraphics[width=\linewidth]{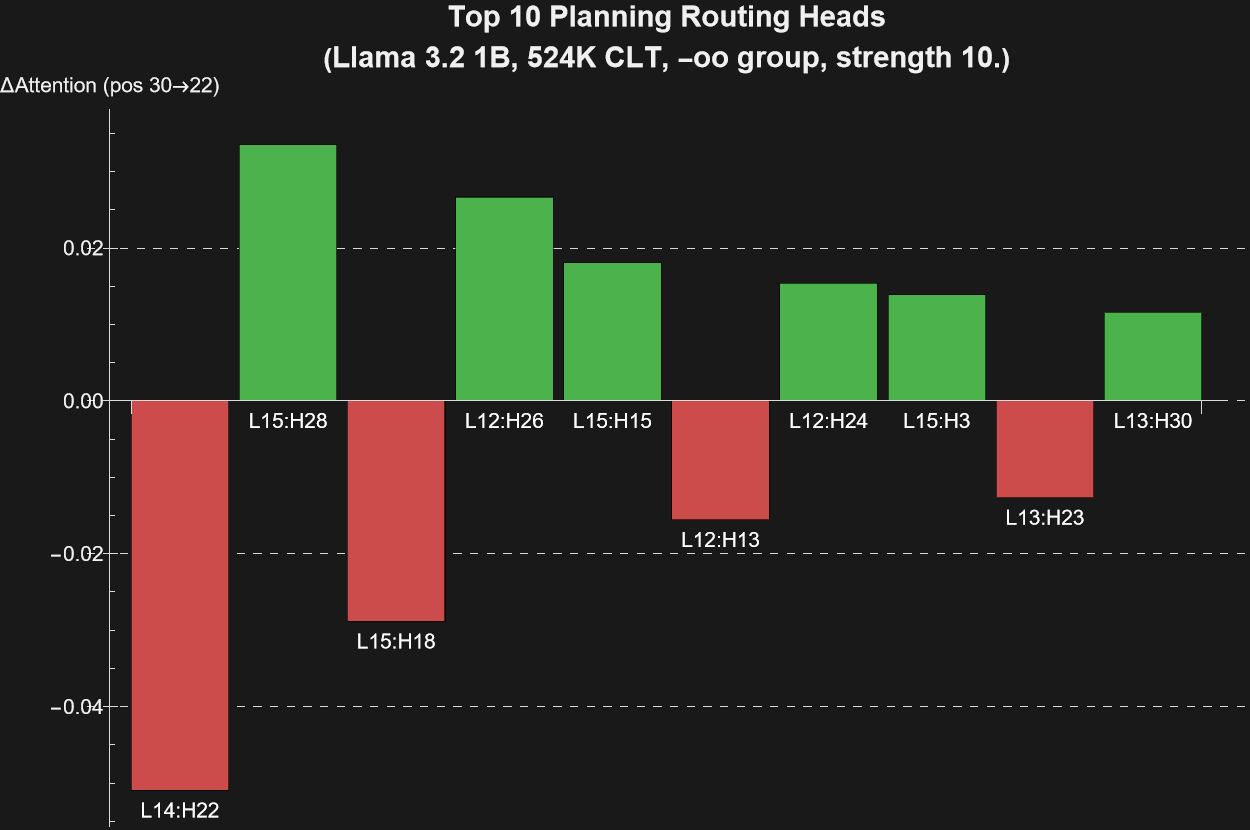}
\smallskip
{\scriptsize (d) Llama -oo}
\end{minipage}
\hfill
\begin{minipage}{0.19\linewidth}
\centering
\includegraphics[width=\linewidth]{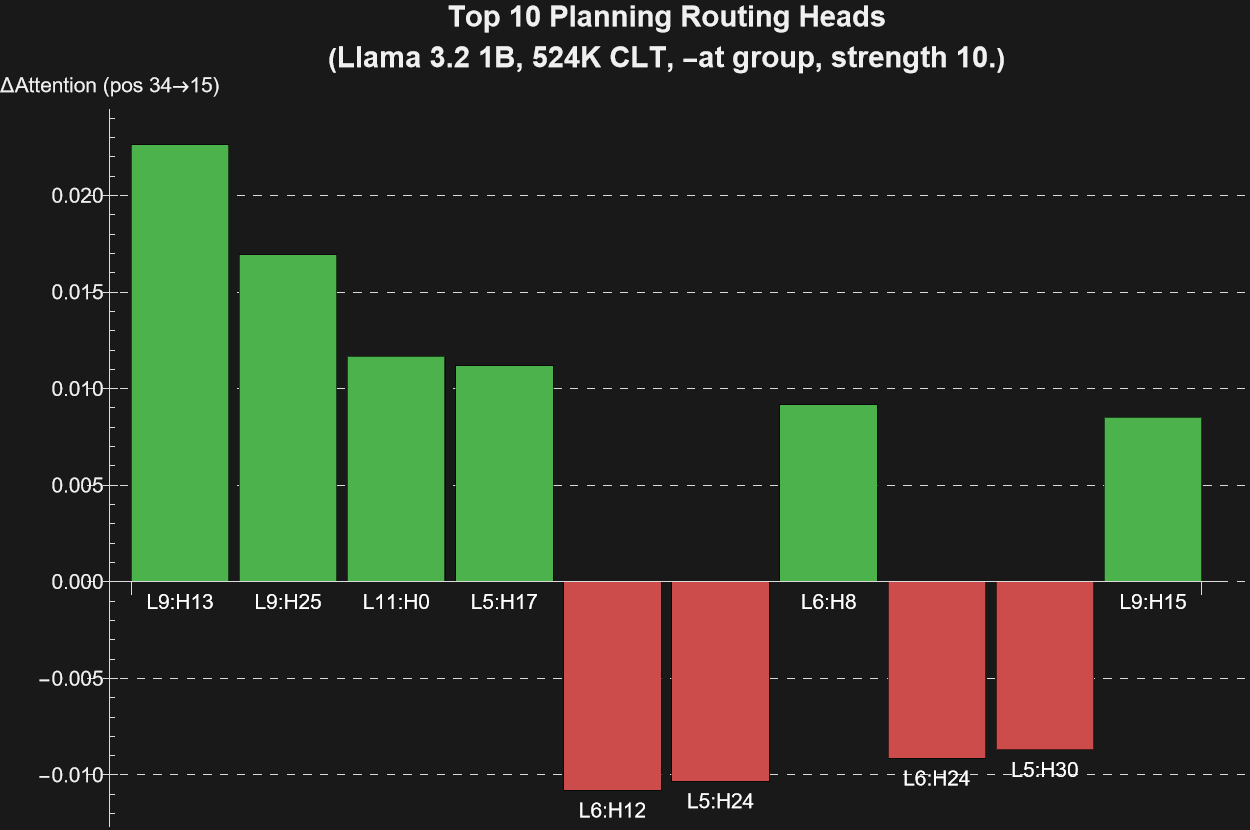}
\smallskip
{\scriptsize (e) Llama -at}
\end{minipage}
\end{center}
\caption{Top 10 routing heads under suppress\,+\,inject. Green: increased attention to the planning site; red: disengagement. (a)~Gemma: L21:H5 dominates consistently. (b--e)~Llama, 4 prompts across 3 rhyme groups: each prompt recruits a different dominant head. The recurring planning heads (2+ prompts) have zero overlap with the factual routing top-10 (Table~\ref{tab:overlap}).}
\label{fig:routing}
\end{figure}

%\smallskip
\paragraph{Attention routing heads.} We define an \emph{attention routing head} as a head whose attention pattern from the output position to the planning-site position changes significantly under suppress\,+\,inject intervention. Three such heads dominate (Figure~\ref{fig:routing}a): \textbf{L21:H5} (layer~21, head~5) ($\Delta = -0.046$, disengages from the natural rhyme plan), L20:H6 ($+0.027$, picks up the injected signal), and L17:H4 ($+0.014$, earlier routing head). The pattern is a push-pull redistribution: some heads disengage from the planning site while others engage. H5 appears at multiple layers (L21, L23, L24, L25) across both CLT granularities, suggesting a structurally dedicated routing channel.

%\smallskip
\paragraph{Suppress amplification.} Inject-only produces $13\times$ weaker routing than suppress\,+\,inject on the 426K CLT: L21:H5 delta is $-0.0035$ (inject only) vs.\ $-0.046$ (suppress\,+\,inject); total routing is 0.017 vs.\ 0.191. L21:H5 also dominates at the 2.5M scale ($\Delta = -0.009$), confirming the effect across CLT granularities; the smaller magnitude reflects the 2.5M inject feature being at L25 (one downstream layer vs.\ four for the 426K feature at L22). The $13\times$ ratio is from a single Gemma prompt; we did not measure it across all 136 pairs or on Llama. Suppress clears the channel; inject fills it.

%\smallskip
\paragraph{Two attractor regimes.} Sweeping intervention strength from 0 to 20 reveals qualitatively different profiles. Factual recall has a hard attractor basin: below ${\sim}1.2\times$ the contrastive distance, the residual stream snaps back within 1--2 layers. Planning has no such basin; perturbations produce proportional routing changes until saturation at ${\sim}15\times$. These are fundamentally different computational substrates within the same model, requiring different measurement tools.

\begin{figure}
\begin{center}
\begin{minipage}{0.48\linewidth}
\centering
\includegraphics[width=\linewidth]{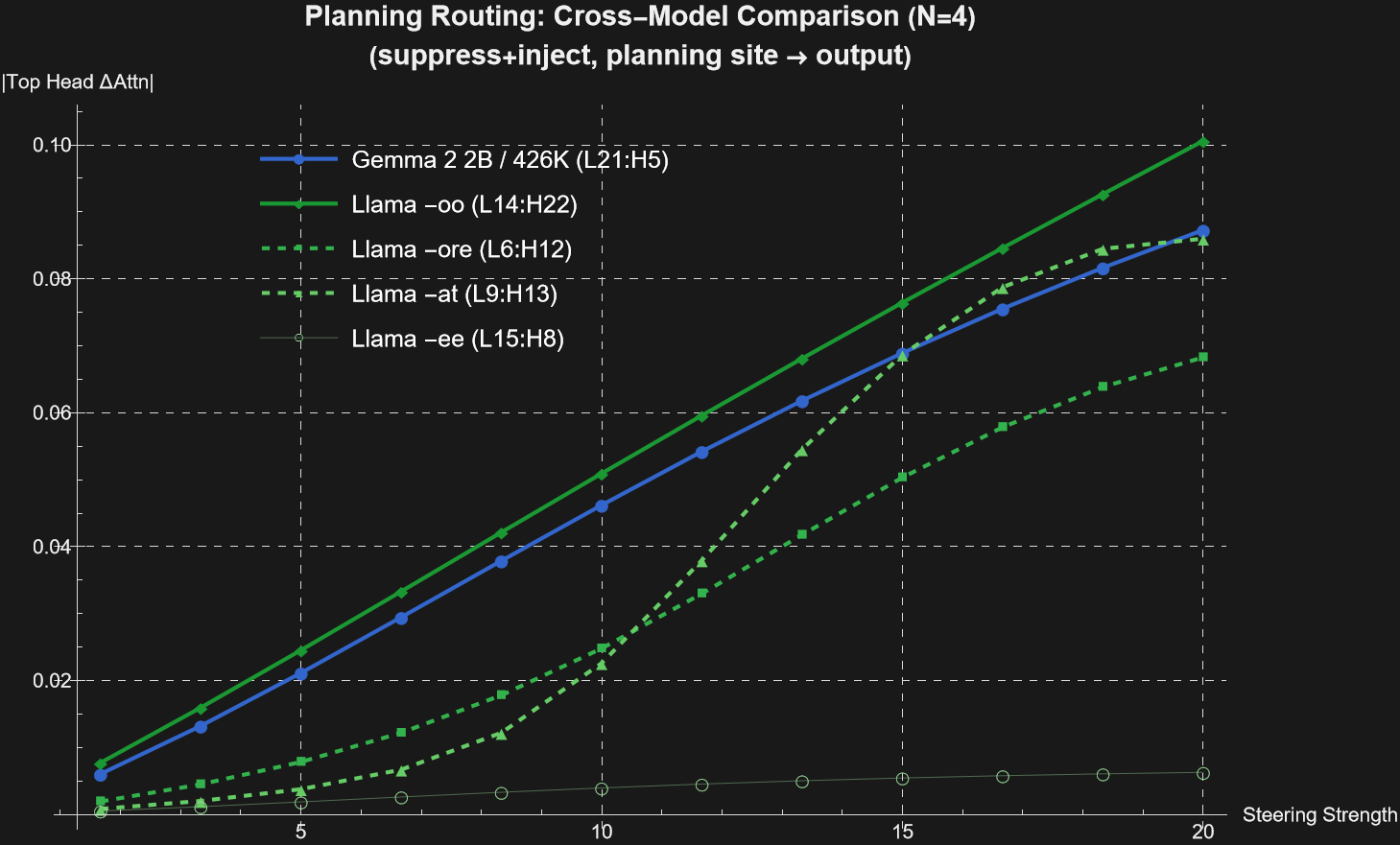}
\smallskip
(a) Top head $|\Delta|$ vs.\ strength
\end{minipage}
\hfill
\begin{minipage}{0.48\linewidth}
\centering
\includegraphics[width=\linewidth]{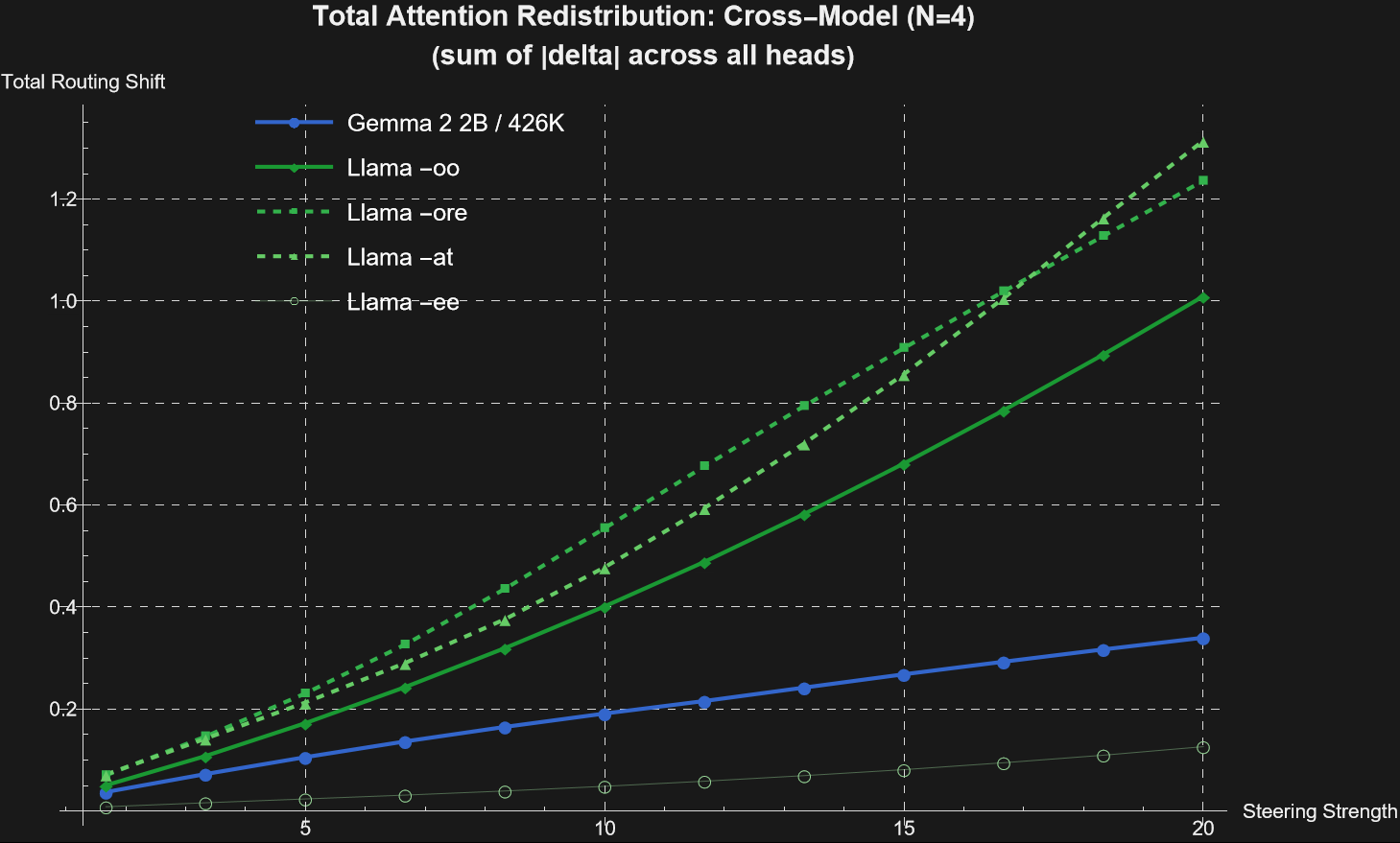}
\smallskip
(b) Total routing shift vs.\ strength
\end{minipage}
\end{center}
\caption{Cross-model strength sweep (Gemma blue, four Llama prompts green). (a)~Top head delta: the strongest Llama prompt (-oo, L14:H22) exceeds Gemma at high strengths, but others (-ee) are much weaker. All show soft, roughly linear responses with no hard threshold. (b)~Total attention redistribution: all four Llama prompts exceed Gemma despite having no dominant head. Llama compensates with distributed routing, while Gemma concentrates the signal through L21:H5.}
\label{fig:strength}
\end{figure}

\paragraph{A3.} Through specific attention routing heads, led by L21:H5, via a push-pull redistribution of attention patterns. The mechanism requires suppress\,+\,inject ($13\times$ amplification) and operates through a soft attractor regime distinct from factual recall (Figure~\ref{fig:strength}). This directly fills the gap \citet{lindsey2025biology} flagged: ``One crucial interaction (seems) to be mediated by changing where attention heads attend\ldots\ \textbf{This is invisible to our current approach}.'' But if attention routing carries the decision, why does Llama, which also activates rhyme features, fail to produce reliable rhymes?

%% ============================================================
\section{Q4: What is the minimum depth for the planning circuit?}
\label{sec:commitment}

\paragraph{Both models search; only one commits.} Llama~3.2 1B knows phonological neighborhoods: its CLT features produce 77.7\% redirect when externally injected. The logit lens (projecting intermediate residual stream states to the vocabulary; \citealp{nostalgebraist2020logit}) shows rhyme words appearing in intermediate layers (L3--L14). Yet Llama produces rhyming couplets unreliably.

The answer is architectural depth. In Gemma, rhyme words appear at L14 and persist through L25 (the output). In Llama, they appear transiently (L11--L14) then dissipate; the signal is absent at L15. The CLT feature distribution confirms: Gemma is bimodal (48.5\% at L0--L1, 38\% at L24--L25), while Llama crams 82\% into L15.

%\smallskip
\paragraph{Causal evidence: layer suppression.} We skip different layer groups in Gemma during generation and measure rhyme rate (10 couplets):

\begin{table}
\small
\begin{minipage}{0.52\linewidth}
\centering
\begin{tabular}{lrl}
\toprule
\textbf{Layers skipped} & \textbf{Rate} & \textbf{Finding} \\
\midrule
None (baseline) & 2/10 & Format pred.\ dominates \\
L5--L9 & 7/10 & Removing it helps \\
L22--L25 & \textbf{0/10} & \textbf{Commitment necessary} \\
L24--L25 & 8/10 & Removing it helps \\
\bottomrule
\end{tabular}
\smallskip
\centerline{(a) Layer suppression, Gemma~2 2B}
\end{minipage}
\hfill
\begin{minipage}{0.44\linewidth}
\centering
\begin{tabular}{lrr}
\toprule
\textbf{Block} & \textbf{Changes} & \textbf{Rate} \\
\midrule
{[0--3]} & 134/596 & 22.5\% \\
{[4--7]} & 124/596 & 20.8\% \\
{[8--11]} & 105/596 & 17.6\% \\
{[12--15]} & \textbf{0/596} & \textbf{0.0\%} \\
\bottomrule
\end{tabular}
\smallskip
\centerline{(b) CF patching, Llama~3.2 1B}
\end{minipage}
\caption{Causal evidence for early commitment. (a)~Skipping L22--L25 in Gemma destroys rhyming. (b)~Counterfactual patching at blocks [12--15] produces zero fact-token changes in Llama.}
\label{tab:causal}
\end{table}

The 0/10 result at L22--L25 gives a one-sided 95\% CI of [0\%, 26\%] (Clopper--Pearson); the 0/596 result at blocks [12--15] gives [0\%, 0.6\%]. Both are consistent with a true rate of zero. Three competing computations: format prediction (L5--L9), rhyme planning (L10--L22), and output refinement (L24--L25). We did not test L10--L16 or L17--L21 in isolation; the three-stage decomposition is based on the four conditions shown and should be considered preliminary.

%\smallskip
\paragraph{The two-stage model.} Both models search. Only Gemma commits. Llama's total attention redistribution under suppress\,+\,inject actually exceeds Gemma's (Figure~\ref{fig:strength}b), but the signal is spread across many heads rather than channeled through a dominant routing head. This distributed routing may be insufficient to sustain a coherent decision across layers. On this two-model comparison, search fits within ${\leq}16$ layers while commitment appears only at ${>}16$. The hypothesis this suggests, which depth-matched model families should test, is that the minimum architecture for planning is defined not by parameter count but by layer depth sufficient to sustain a decision from search through to output.

\paragraph{A4.} The evidence is consistent with search within ${\leq}16$ layers and commitment beyond; the boundary tracks depth rather than parameter count in our two-model comparison, a hypothesis rather than an established result. This establishes that the planning circuit has two separable stages, but so far we have only studied rhyme planning. Is the early-commitment, late-routing pattern specific to this one task?

%% ============================================================
\section{Q5: Is early commitment specific to planning?}
\label{sec:prolepsis}

The previous sections establish a structural motif for rhyme planning: early feature activation, attention routing, no correction. We now test whether the same motif appears in a different task, factual recall, on Llama~3.2 1B (the Transluce dataset \citep{li2025transluce} was generated for the Llama family; extending to Gemma would require re-screening CounterFact pairs, which we leave to future work).

%\smallskip
\paragraph{Early commitment in factual recall.} Using 89 gold pairs from the CounterFact dataset \citep{meng2022locating,li2025transluce} on Llama~3.2 1B, we replace the residual stream at subject-token positions with the counterfactual's residual stream, sweeping across 4-layer blocks. Fact tokens are $4.5\times$ more causally effective than template tokens (15.2\% vs.\ 3.4\%). The block-by-block gradient (Table~\ref{tab:causal}b) shows a monotonic decline: 22.5\% at blocks [0--3] down to 0.0\% at blocks [12--15]. The factual decision is irrevocable by layer~12. Of 664 changed predictions, 48.3\% flip to the exact counterfactual target; the intervention redirects, not merely disrupts. The remaining 51.7\% shift to other tokens, typically semantically related to the counterfactual subject (e.g., other car manufacturers when patching ``Ferrari''), consistent with partial activation of the counterfactual neighborhood rather than random disruption. KL divergence cleanly separates changed from unchanged patches ($80\times$: mean 0.438 vs.\ 0.005).

%\smallskip
\paragraph{Factual routing heads.} We measure per-head attention deltas when patching subject tokens. L15:H8 dominates at 48.6\% appearance rate ($2.5\times$ the next head), with 79.8\% negative direction: when the wrong subject is injected, it \emph{reduces} attention to the original fact. The routing shift confirms the causal gradient: 0.384 at blocks [0--3], declining to 0.063 at blocks [12--15].

%\smallskip
\paragraph{Zero cross-task head overlap.}

\begin{table}
\small
\begin{center}
\begin{tabular}{lcc}
\toprule
\textbf{Property} & \textbf{Planning (4 prompts)} & \textbf{Factual recall (89 pairs)} \\
\midrule
Dominant head & Prompt-specific & L15:H8 (48.6\%) \\
Recurring heads (2+) & L5:H17, L6:H12, L6:H24, L9:H13, L11:H0 & L15:H8, L14:H27, L15:H16 \\
Mean layer (recurring) & 7.4 (46\% depth) & 12.8 (80\% depth) \\
Recurring $\cap$ factual top-10 & \multicolumn{2}{c}{\textbf{0 / 5}} \\
\bottomrule
\end{tabular}
\end{center}
\caption{Cross-task comparison on Llama~3.2 1B. Planning routing recruits prompt-specific mid-layer heads; factual routing recruits consistent late-layer heads. The recurring planning heads have zero overlap with the factual top-10.}
\label{tab:overlap}
\end{table}

Planning routing is prompt-specific: each rhyme group recruits a different dominant head (L15:H8 for -ee, L6:H12 for -ore, L14:H22 for -oo, L9:H13 for -at). Notably, the -ee prompt's dominant head (L15:H8) is also the dominant factual routing head; however, L15:H8 does not recur in any other planning prompt, indicating it is a general-purpose late-layer head co-opted by this particular prompt rather than a dedicated planning head. The heads that recur across 2+ prompts (L6:H12, L9:H13, L5:H17, L11:H0, L6:H24) are concentrated in mid-layers (5--11), while the factual routing heads are concentrated in late layers (9--15). The recurring planning heads have zero overlap with the factual top-10. Per-prompt top-10 lists show 0--3 heads in common with the factual top-10, but these shared heads do not recur across prompts, suggesting the overlap is incidental rather than structural (though more prompts would be needed to confirm this).

%\smallskip
\paragraph{Prolepsis as a structural motif.} We term this shared pattern \textbf{prolepsis}: the structural motif in which a transformer commits to a decision at early layers, and attention routing heads sustain that commitment without correction, even when the committed answer is wrong. The identification of recurring structural motifs across tasks echoes prior work on grokking circuits \citep{nanda2023progress} and sparse feature circuits \citep{marks2024sparse}, but prolepsis is distinguished by its cross-task invariance: the \emph{template} is shared while the \emph{substrates} differ.

The zero aggregate overlap, combined with the layer-depth separation (mid-layer for planning, late-layer for factual recall), shows that prolepsis is not implemented by a fixed set of ``commitment heads.'' Different tasks recruit heads at different network depths, but the template (early commitment, routing-mediated propagation, zero correction) is shared. \textbf{For these two tasks, prolepsis is in the architecture, not in the wiring.}

Sustained propagation is the non-obvious finding. Early commitment alone is unsurprising for retrieval. The striking result is that nothing corrects it through 75\% of the remaining depth. Routing heads respond to every perturbation (they are not idle) but sustain rather than reverse the commitment.

\paragraph{A5.} No. Prolepsis is not specific to planning. The same structural motif (early commitment, routing-mediated propagation, zero correction) appears in factual recall with routing at a different network depth and through different heads. In the decoder-only models tested, prolepsis recurs across tasks.

%% ============================================================
\section{Discussion and conclusion}
\label{sec:discussion}

\paragraph{Prolepsis recurs across tasks and scales.} Two models tested by us, consistent with Anthropic's published results on a third (Claude 3.5 Haiku); two organizations; two tasks; three CLT scales; four planning prompts across three rhyme groups: the same pattern. This is not a training artifact.

\paragraph{Planning requires CLTs to observe.} Six failed steering methods and ${<}0.5\%$ redirect rates establish that planning occupies a different computational substrate than factual recall. Tools developed for one regime may be blind to the other. This substrate distinction may run deeper: prior work localizes factual knowledge in MLP layers \citep{meng2022locating,dai2022knowledge}, while our results localize planning routing in attention heads. If factual recall operates primarily through MLPs and planning through attention, prolepsis encompasses two mechanistically distinct circuits that share only the template (early commitment, sustained propagation, zero correction). We did not measure MLP contributions to planning; this dissociation remains a hypothesis.

\paragraph{Depth, not parameters, determines commitment.} Llama (16 layers) searches but cannot commit. Gemma (26 layers) can. The minimum architecture requires sufficient layer depth, not parameter count. We note that Llama and Gemma also differ in training data, tokenizer, and architectural details (GQA, sliding window attention), so the depth claim is suggestive rather than causal. Isolating depth would require a model family with varying depths but shared training (e.g., Pythia); this is blocked by the absence of CLTs for such families.

\paragraph{Is irrevocability specific to prolepsis?} An alternative explanation is that autoregressive transformers generally do not reverse concentrated probability mass: pre-norm LayerNorm and the additive residual connection preserve dominant directions, and subsequent layers refine rather than reverse. Under this view, irrevocability is a generic property of residual streams, not circuit-specific. We cannot rule this out; however, the task-specific routing heads at different depths suggest irrevocability is at least \emph{maintained} by dedicated circuitry. Prolepsis differs from ``early exiting'' \citep{schwartz2020right}, where models skip layers for efficiency: here, all layers compute but none corrects. We tested this directly: injecting a contradictory feature at post-commitment layers (L23--L25) at up to $2\times$ the commitment strength fails to change the L21:H5 routing delta or redirect the output; P(commit word) \emph{increases} while P(correct word) remains negligible (Appendix~\ref{app:correction}). The commitment is irrevocable, which has implications for safety: a model that irrevocably commits at early layers is harder to redirect than one that deliberates across its full depth.

\paragraph{Limitations and CLT coverage.} Our models constitute the complete set with open CLTs on consumer hardware. Both tasks are retrieval-like; whether prolepsis holds for deliberation is unknown: we exhaustively scanned 2.5M features for code and cooking domains and found zero task-specific features (Appendix~\ref{app:domains}). The Gemma attention routing deltas are from a single prompt per CLT scale; L21:H5 dominates in both, but we do not report variance across the full set of 136 pairs. The planning routing on Llama uses 4 prompts across 3 rhyme groups (vs.\ 89 pairs for factual recall); each prompt recruits different heads, so the planning routing signal is less stable than the factual routing signal. Planning-site localization requires CLT features that discretely encode the constraint domain. Our exhaustive scan of 2.5M features (Appendix~\ref{app:domains}) reveals that CLTs at current scales develop discrete features only for high-frequency, phonologically distinctive domains (rhyme groups from web poetry); code types and culinary semantics produce zero domain-specific features despite being present in the training data. This CLT coverage boundary defines the current limits of the methodology and has implications beyond this paper.

\paragraph{Conclusion.} Transformers implement prolepsis: early irrevocable commitment sustained through attention routing at task-dependent network depths. The motif manifests in both rhyme planning (mid-layer routing) and factual recall (late-layer routing), across architectures and scales, with zero overlap between recurring planning heads and the factual top-10. The evidence is consistent with a minimum architecture defined by layer depth: search within ${\leq}16$ layers, commitment beyond, a hypothesis awaiting depth-matched model families. The planning circuit is invisible to the six residual stream methods we tested; among them, only CLT-based steering observes it. All experiments reproduce on a single consumer GPU.

%% ============================================================

\section*{Reproducibility statement}

All experiments were conducted on a single consumer GPU (16\,GB VRAM) using a custom open-source framework, open-weights models (Gemma~2 2B, Llama~3.2 1B), and open CLTs hosted on HuggingFace. The complete set of commands to reproduce every experiment is provided in Appendix~\ref{app:reproduction}. Our CLT encoder/decoder was validated against the Python Circuit Tracer reference implementation: 90/90 top-10 features match with maximum relative error $1.2 \times 10^{-6}$. The CounterFact gold pairs were derived from a publicly available dataset \citep{li2025transluce}; our screening criteria and the resulting 89 pairs are fully specified in Section~\ref{sec:setup}. Code and data will be released upon publication.

\section*{LLM usage disclosure}

An LLM (Claude, Anthropic) was used as a research assistant throughout this work: paper structuring, drafting and revising text, experimental design (including the correction test in Appendix~\ref{app:correction}), code generation for experimental tooling, data analysis and interpretation, consistency checking, and literature review. All experimental results were produced by independently built, deterministic software on the author's hardware. All LLM contributions were verified by the author, who takes full responsibility for the contents of this submission.

\bibliography{references}
\bibliographystyle{colm2026_conference}

\appendix

\section*{Appendix overview}

This appendix provides full experimental details, data tables, and reproduction commands for all results in the main text. All tables and statistics were generated automatically from the experiment output files (JSON) using scripts; no numbers were transcribed manually.

\smallskip
\noindent
\small
\begin{tabular}{@{}lp{0.72\linewidth}@{}}
\toprule
\textbf{Appendix} & \textbf{Contents} \\
\midrule
\hyperref[app:steering]{A. Steering method details} & Six failed residual-stream methods, root cause analysis, two key insights \\
\hyperref[app:discovery]{B. CLT feature discovery protocol} & Vocabulary scan, CMU dictionary cross-reference, 35 Gemma + 12 Llama rhyme groups, cross-model comparison \\
\hyperref[app:domains]{C. Domain generalization} & Exhaustive scan of 2.5M features for code (1{,}883 hits, all false positives) and cooking (628 hits, all false positives) \\
\hyperref[app:sweeps]{D. Full position sweep data} & 136 + 264 + 47 pairs across 3 CLT configurations, P(inject) distributions, cross-model summary \\
\hyperref[app:heads]{E. Attention routing head tables} & Top-20 heads for Gemma (2 CLTs) and Llama (5 prompts), recurring heads, cross-task overlap verification \\
\hyperref[app:reproduction]{F. Reproduction commands} & Complete commands for all experiments, hardware and runtime table \\
\hyperref[app:correction]{G. Direct test of irrevocability} & Contradictory injection at post-commitment layers; correction sweep 0--20; L21:H5 locked \\
\bottomrule
\end{tabular}
\normalsize
\bigskip

\section{Steering method details}
\label{app:steering}

Before discovering the bottom-up CLT approach (Section~\ref{sec:setup}), we applied six top-down methods to redirect rhyme endings in Gemma~2 2B. All six fail completely (0\% target hit rate). This appendix documents each method and diagnoses the root cause.

\subsection{Method 1: Max-activation probes}

For each CLT feature, we find the token position where it activates most strongly during a forward pass on a rhyming prompt. We then inject that feature at the planning site and check whether the model's top-1 prediction changes to the target rhyme word.

\textbf{Result:} 0\% target hit across all 426K features. The features that activate most strongly at the planning site encode semantic neighborhoods (e.g., ``outdoor/nature'' concepts), not individual rhyme words. No single feature is specific enough to select a particular word.

\subsection{Method 2a/b: Decoder dot product and cosine similarity}

We compute the dot product (2a) and cosine similarity (2b) between each feature's decoder vector and the target word's embedding \citep{nostalgebraist2020logit}. Features with the highest scores are injected at the planning site.

\textbf{Result:} 0\% target hit. Method 2a produces catastrophic degradation at high strengths (the model outputs garbage), because high-dot-product features tend to have large norms that overwhelm the residual stream. Method 2b avoids the norm issue but selects features that encode the target word's semantic neighborhood rather than the word itself.

\subsection{Method 3: Planning-site activation + cosine filter}

Combine Methods 1 and 2: first filter features by cosine similarity to the target word, then rank by activation strength at the planning site.

\textbf{Result:} 0\% target hit. The intersection of ``activates at the planning site'' and ``projects toward the target word'' is empty or near-empty for all target words tested.

\subsection{Method 4: Method 3 + multi-layer clamping}

Extend Method 3 by clamping the selected features across multiple consecutive layers (not just one), hoping to sustain the signal.

\textbf{Result:} 0\% target hit. Multi-layer clamping increases the intervention's magnitude but does not change the feature's semantic content. Clamping a ``nature/outdoor'' feature across 4 layers still produces ``nature/outdoor'' output, not a specific rhyme word.

\subsection{Method 5: Contrastive word probes}

For each rhyme group pair (e.g., ``-air'' vs.\ ``-out''), compute contrastive feature vectors (mean activation difference between prompts ending in each group), following the activation addition paradigm \citep{turner2023activation,zou2023representation}. Inject the contrastive vector at the planning site.

\textbf{Result:} 0\% target hit across 120 pairs, 7 strength levels (0--16), 3 samples each (2,520 total evaluations). The contrastive vectors capture the distributional difference between rhyme groups but cannot redirect the model's top-1 prediction. Even at strength 16, target hit rate remains 0\%.

\subsection{Method 6: Causal activation patching}

Replace the residual stream at the planning-site position with the residual stream from a different prompt (one that would naturally produce the target rhyme), following the causal mediation framework \citep{vig2020investigating,geiger2021causal}. This is the strongest possible intervention: it replaces the entire activation, not just a single feature.

\textbf{Result:} The intervention removes 9.72 logit units from the natural top prediction (massive causal effect), but 0\% target hit. The patched activation disrupts the model's natural prediction without redirecting it to the intended target.

\subsection{Root cause analysis}

The logit gap between the model's natural top prediction and any target rhyme word is ${\sim}25$ logit units. With 426K features across 26 layers (${\sim}16$K per layer) covering a 256K vocabulary, no small group of features is specific enough to act as a lexical selector. The features capture semantic neighborhoods, not individual words. At the 426K scale, rhyme planning operates at the \emph{rhyme-group} level: features distinguish ``-air'' from ``-out'' but not ``chair'' from ``stair.''

Two key insights emerged from these failures:
\begin{enumerate}
    \item \textbf{Binary argmax is the wrong metric.} All six methods measure whether the target word becomes the model's top-1 prediction. Anthropic's Figure~13 measures \emph{continuous probability} (P(target)), not top-1 hit. A feature that raises P(target) from $10^{-8}$ to $0.48$ is a strong causal effect even though the target is not top-1.
    \item \textbf{Inject-only is insufficient.} Methods 1--6 all inject features without suppressing the competing natural rhyme group. Anthropic's protocol uses suppress\,+\,inject: first remove the natural signal, then inject the alternative. The $13\times$ suppress amplification effect (Section~\ref{sec:routing}) explains why inject-only fails.
\end{enumerate}

These insights motivated the bottom-up approach described in Section~\ref{sec:setup}: reverse-engineer what CLT features actually encode (vocabulary scan), then apply Anthropic's suppress\,+\,inject protocol with continuous probability measurement.

\section{CLT feature discovery protocol}
\label{app:discovery}

The planning-site localization experiments require CLT features that correspond to specific rhyme groups. Because no catalog of phonologically organized CLT features exists, we built one from scratch via a two-step bottom-up protocol.

\subsection{Step 1: Vocabulary scan}

For each CLT (Gemma 426K, Gemma 2.5M, Llama 524K), we compute the cosine similarity between every feature's decoder vector and every token embedding in the model's vocabulary. For Gemma 426K, this is a $425{,}984 \times 256{,}768$ matmul, chunked into batches of 4{,}096 features to fit 16\,GB VRAM. The scan produces, for each feature, a ranked list of vocabulary tokens by cosine similarity. We retain features whose top-1 token is a clean English word (no subword fragments, no punctuation, no non-English tokens).

\textbf{Yield:} 287 clean English features for Gemma 426K; 79 for Llama 524K. The $3.6\times$ difference reflects the smaller model's sparser feature vocabulary.

\subsection{Step 2: CMU dictionary cross-reference}

We cross-reference the clean English features with the CMU Pronouncing Dictionary \citep{weide1998cmu}, which provides phoneme sequences for ${\sim}134{,}000$ English words. For each feature whose top-1 token appears in the dictionary, we extract the \emph{rhyme ending}: the stressed vowel plus all subsequent phonemes (e.g., ``about'' $\to$ AH0\,B\,AW1\,T $\to$ rhyme ending AW1\,T). Features are grouped by rhyme ending. Groups with ${\geq}2$ words and minimum cosine ${\geq}0.3$ are retained.

\textbf{Yield:} 35 rhyme groups (98 words) for Gemma; 12 rhyme groups (28 words) for Llama.

\subsection{Gemma 2 2B rhyme groups (426K CLT)}

\begin{small}
\begin{tabular}{@{}llp{0.55\linewidth}@{}}
\toprule
\textbf{Ending} & \textbf{$n$} & \textbf{Words (feature ID)} \\
\midrule
IY1 & 8 & we (L25:9391), be (L25:12680), free (L20:12510), he (L22:3693), she (L24:714), me (L21:11371), three (L25:11713), de (L25:14100) \\
OW1 & 7 & so (L25:6778), grow (L25:4985), go (L25:4505), know (L22:10362), slow (L25:5776), though (L20:12770), snow (L19:3248) \\
UW1 & 7 & to (L25:10073), who (L23:1548), do (L25:14014), too (L18:7484), two (L19:5076), new (L23:3304), ou (L25:5927) \\
AW1\,N\,D & 4 & found (L25:6000), round (L24:15683), around (L22:10243), ground (L25:14915) \\
AA1\,N & 4 & don (L25:12569), upon (L24:10328), on (L25:6476), con (L25:5978) \\
AE1\,N & 3 & can (L25:691), an (L25:7363), than (L19:12532) \\
IY1\,Z & 3 & please (L25:1665), these (L19:7913), chinese (L24:1672) \\
AH1\,M & 3 & come (L25:15291), from (L24:11252), become (L22:13319) \\
EY1\,T & 3 & straight (L24:15959), great (L24:2883), weight (L18:6003) \\
AO1\,R & 3 & your (L21:4934), or (L23:7122), for (L24:5705) \\
IH1\,L & 3 & will (L22:9599), still (L19:11977), until (L21:15310) \\
AY1\,ER0 & 3 & require (L24:7275), fire (L22:4814), entire (L20:11687) \\
\midrule
\multicolumn{3}{@{}l}{\emph{Plus 23 groups with 2 words each (e.g., about/out, black/back, at/that, kind/find).}} \\
\multicolumn{3}{@{}l}{\emph{Full list: 35 groups, 98 words total.}} \\
\bottomrule
\end{tabular}
\end{small}

\subsection{Llama 3.2 1B rhyme groups (524K CLT)}

\begin{small}
\begin{tabular}{@{}llp{0.55\linewidth}@{}}
\toprule
\textbf{Ending} & \textbf{$n$} & \textbf{Words (feature ID)} \\
\midrule
IY1 & 4 & he (L13:30985), be (L9:5488), ne (L14:27874), we (L13:32049) \\
AO1\,R & 3 & for (L1:5297), or (L3:22663), more (L10:18203) \\
UW1 & 3 & to (L14:18284), qu (L15:8165), new (L11:20779) \\
AE1\,T & 2 & that (L14:13043), sat (L14:6132) \\
IH1\,Z & 2 & is (L4:8361), his (L3:30599) \\
AA1\,R & 2 & are (L6:14873), ar (L14:1026) \\
ER1 & 2 & were (L4:5192), her (L11:20120) \\
AE1\,N & 2 & an (L14:12231), can (L14:2839) \\
AH0 & 2 & the (L14:7589), le (L14:15075) \\
IH1\,L & 2 & will (L13:26263), il (L15:19385) \\
EY1 & 2 & they (L12:9759), ay (L15:6901) \\
IH1\,SH\,AH0\,N & 2 & addition (L9:2383), nutrition (L3:202) \\
\bottomrule
\end{tabular}
\end{small}

\subsection{Cross-model comparison}

\begin{small}
\begin{tabular}{@{}lcc@{}}
\toprule
\textbf{Metric} & \textbf{Gemma 2 2B} & \textbf{Llama 3.2 1B} \\
\midrule
Total CLT features & 425{,}984 & 524{,}288 \\
Clean English features & 287 & 79 \\
Rhyme groups & 35 & 12 \\
Rhyming words & 98 & 28 \\
Mean words/group & 2.8 & 2.3 \\
Shared rhyme endings & \multicolumn{2}{c}{8 (IY1, AO1\,R, UW1, AE1\,T, AA1\,R, AE1\,N, IH1\,L, EY1)} \\
\bottomrule
\end{tabular}
\end{small}

\smallskip
\noindent Despite having more total features (524K vs.\ 426K), Llama produces $3.6\times$ fewer clean English features and $2.9\times$ fewer rhyme groups. This reflects the smaller model's training distribution: with fewer parameters, fewer features develop word-level specificity. The 8 shared rhyme endings confirm that both models learn phonological structure from overlapping training data, but at different granularities.

\subsection{Observation: features cluster by phonetic ending, not semantics}

The vocabulary scan reveals that CLT features at the 426K scale encode \emph{phonological neighborhoods}, not semantic categories. For example, the AW1\,N\,D group (found, round, around, ground) clusters words that share no semantic relationship but share a phonetic ending. Similarly, the IY1 group (we, be, free, he, she, me, three, de) spans pronouns, verbs, adjectives, and numbers. This phonological organization is what makes planning-site localization possible: the model has features that respond to ``sounds like X,'' which is exactly the computation needed for rhyme planning.

\section{Domain generalization: code and cooking}
\label{app:domains}

We systematically tested whether planning-site localization could extend to two alternative constraint domains: code generation (where a function signature constrains the return type) and cooking (where a dish name constrains ingredients and preparation steps). In both cases, the prerequisite is CLT features that discretely encode the constraint vocabulary. We scanned all 2{,}555{,}904 features of the 2.5M Gemma CLT (and confirmed the pattern at 426K scale) for domain-specific tokens.

\subsection{Code domain: zero type-specific features}

We searched for 40 programming type tokens (\texttt{int}, \texttt{str}, \texttt{float}, \texttt{bool}, \texttt{list}, \texttt{dict}, \texttt{void}, \texttt{String}, \texttt{Array}, \texttt{long}, \texttt{short}, \texttt{set}, etc.) and 20 control-flow keywords (\texttt{for}, \texttt{if}, \texttt{return}, \texttt{case}, \texttt{throw}, etc.).

\textbf{Results at 2.5M scale:} 1{,}883 apparent hits (80 type names, 310 control-flow, 482 common identifiers, 907 delimiters, 104 structural). At 426K: 575 hits (30 type, 106 control-flow).

\textbf{Every type-name hit is an English adjective or verb:}
\begin{itemize}
    \item \texttt{long} (38 of 80 hits): top-5 decoder contexts are ``long,'' ``Long,'' ``LONG,'' ``dolgo'' (Russian for ``long''), ``longa'' (Portuguese). These are natural-language duration features.
    \item \texttt{short} (4 hits): English adjective for brevity.
    \item \texttt{set} (10 hits): English verb ``to set,'' with decoder contexts ``set,'' ``Set,'' ``sets,'' ``SET.''
\end{itemize}
Zero features encode \texttt{int}, \texttt{str}, \texttt{float}, \texttt{bool}, \texttt{dict}, \texttt{void}, \texttt{String}, or \texttt{Array}.

\textbf{Every control-flow hit is an English preposition, conjunction, or verb:}
\begin{itemize}
    \item \texttt{for} (155 of 310 hits): decoder contexts include translations of the English preposition into Russian (``dlya''), Farsi (``baray''), Greek (``gia''), German (``f\"ur''), Indonesian (``untuk'').
    \item \texttt{if} (14 hits): decoder contexts include ``yesli'' (Russian), ``ruguo'' (Chinese), ``jika'' (Indonesian). The English conjunction, not the conditional keyword.
    \item \texttt{return} / \texttt{throw}: decoder contexts show verb conjugations (``returned,'' ``thrown,'' ``threw''), not code statements.
\end{itemize}

Even at 98{,}304 features per layer, the CLT does not develop code-specific representations because code is a small fraction of Gemma~2 2B's training corpus.

\subsection{Cooking domain: zero culinary-specific features}

We searched for 170+ culinary keywords across six categories: cooking verbs (beat, cut, freeze, mix, reduce, rest, serve, smoke, stuff), descriptors (brown, cold, dark, done, golden, ground, hard, hot, large, raw, round, small, sweet, thin, thick, warm, whole), ingredients (butter, chicken, egg, fish, meat, milk, oil, salt, sugar, water, wine), measures (fast, half, quart, quick, slow), dish types (chili, spread, stock), and tools (oven, pot, sheet, wrap).

\textbf{Results:} 628 apparent hits (349 descriptors, 122 cooking verbs, 117 ingredients, 27 measures, 7 dish types, 6 tools).

\textbf{Every hit is a common English word with no culinary specificity:}
\begin{itemize}
    \item ``rest'' (62 of 122 cooking-verb hits): decoder contexts are ``rest,'' ``Rest,'' ``REST,'' ``resta'' (Spanish). The English verb ``to relax,'' not ``let the meat rest.''
    \item ``ground'' (39 of 349 descriptor hits): \emph{the same features already in our rhyme groups}. The English adjective as in ``ground floor,'' not ``ground beef.''
    \item ``water'' (37 of 117 ingredient hits): general substance, decoder contexts include the Chinese character for water. Not recipe-specific.
    \item ``cut'' (24 hits): generic English verb for severing, not a culinary technique.
    \item ``stock'' (dish type): decoder top-5 includes ``market.'' Financial stock, not cooking stock.
\end{itemize}

The CLT has no features that distinguish ``ground beef'' from ``ground floor'' or ``reduce the sauce'' from ``reduce emissions.'' The training distribution does not produce discrete culinary feature clusters at this scale.

\subsection{Why rhyming works and appears to be unique}

Rhyme groups earned dedicated CLT features because web poetry and nursery rhymes are frequent enough in the training distribution to warrant specialized phonological representations. Direct evidence: wrapping rhyme-setting words in \texttt{<strong>} HTML tags increases the rhyme rate from 78\% to 81\% (Section~\ref{sec:setup}), indicating the model learned from HTML-formatted web poetry.

Code types and culinary semantics, while present in the training data, do not produce discrete, high-frequency feature clusters at either CLT scale. This does not imply that the models cannot plan in these domains; it implies that planning-site localization, which requires discrete domain-specific CLT features, cannot be applied. The models may plan for code or cooking through mechanisms that are not decomposable into the kind of phonological clusters that make rhyme planning visible. The absence of alternative testbeds for planning-site localization is the result of this exhaustive search across the available feature space, not a failure to look.

\section{Full position sweep data}
\label{app:sweeps}

We report summary statistics for all suppress\,+\,inject position sweeps across the three CLT configurations. In each sweep, we suppress the natural rhyme group's CLT features and inject an alternative feature at every token position, measuring P(injected word) at each position. A pair ``localizes'' if the maximum P(injected word) occurs at the planning site (the last token position).

\subsection{Gemma 2 2B, 426K CLT: 136 pairs}

4 prompts $\times$ 34 alternative rhyme groups. Three natural groups: ``-out'' (2 prompts, 68 pairs), ``-ow'' (1 prompt, 34 pairs), ``-oo'' (1 prompt, 34 pairs). Strength: 10.0.

\textbf{Localization rate:} 95/136 = 69.9\%. The 41 non-localizing pairs typically have very weak redirects ($P < 10^{-6}$) where noise dominates.

\begin{small}
\begin{tabular}{@{}lrrrrr@{}}
\toprule
\textbf{P(inject) bin} & ${\geq}0.1$ & $[0.01, 0.1)$ & $[10^{-3}, 0.01)$ & $[10^{-6}, 10^{-3})$ & ${<}10^{-6}$ \\
\midrule
Count & 9 & 3 & 1 & 25 & 98 \\
\bottomrule
\end{tabular}
\end{small}

\smallskip
\noindent\textbf{Best redirects:} P(``at'') = 0.986 ($2.1 \times 10^{9}\times$ ratio); P(``around'') = 0.483 ($10^{7}\times$). \textbf{Weakest:} P(``city'') = $2.4 \times 10^{-14}$ ($86\times$ ratio, peak not at planning site).

The 9 pairs with P ${\geq}$ 0.1 all come from high-cosine rhyme groups (``-at,'' ``-ound,'' ``-ack'') where the CLT features are most word-specific. The 98 pairs with P ${<} 10^{-6}$ come from groups with low cosine scores or function-word features (``this,'' ``city,'' ``pretty''), where the CLT encodes a broad semantic neighborhood rather than a phonological target.

\subsection{Gemma 2 2B, 2.5M CLT: 264 pairs}

4 prompts $\times$ 66 alternative groups (word-level features at 98{,}304 per layer). Same 3 natural groups. Strength: 10.0.

\textbf{Localization rate:} 164/264 = 62.1\%.

\begin{small}
\begin{tabular}{@{}lrrrrr@{}}
\toprule
\textbf{P(inject) bin} & ${\geq}0.1$ & $[0.01, 0.1)$ & $[10^{-3}, 0.01)$ & $[10^{-6}, 10^{-3})$ & ${<}10^{-6}$ \\
\midrule
Count & 16 & 5 & 3 & 35 & 205 \\
\bottomrule
\end{tabular}
\end{small}

\smallskip
\noindent\textbf{Best redirect:} P(``black'') = 0.522 ($3.4 \times 10^{11}\times$). \textbf{Best ratio:} P(``kind'') at $3.78 \times 10^{12}\times$. At 2.5M, every rhyme word has its own dedicated feature (209 words, mean decoder rank 1.0), but many produce weak redirects because the inject feature's decoder vector has low norm at the output layer.

67 unique inject words, all individual English words (not groups). The 16 pairs with P ${\geq}$ 0.1 include ``black,'' ``can,'' ``kind,'' ``well,'' ``that,'' ``round,'' confirming that word-level features can produce strong redirects when the feature is high-cosine and phonologically specific.

\subsection{Llama 3.2 1B, 524K CLT: 47 pairs}

4 prompts $\times$ 11--12 alternative groups. Three natural groups: ``-ee'' (2 prompts, 24 pairs), ``-ore'' (1 prompt, 12 pairs), ``-oo'' (1 prompt, 11 pairs). Strength: 15.0.

\textbf{Localization rate:} 40/47 = 85.1\%. Only 7 pairs fail to localize (all involving weak features: ``new,'' ``the'').

\begin{small}
\begin{tabular}{@{}lrrrrr@{}}
\toprule
\textbf{P(inject) bin} & ${\geq}0.1$ & $[0.01, 0.1)$ & $[10^{-3}, 0.01)$ & $[10^{-6}, 10^{-3})$ & ${<}10^{-6}$ \\
\midrule
Count & 4 & 0 & 6 & 27 & 10 \\
\bottomrule
\end{tabular}
\end{small}

\smallskip
\noindent\textbf{Best redirect (initial 47-pair sweep):} P(``that'') = 0.834 ($21{,}829\times$). A re-run with relabeled rhyme groups covers 44 pairs (4 prompts $\times$ 11 groups) and localizes 36/44 (81.8\%) with best P(``that'') = 0.777, the values quoted in the main text; the two runs are otherwise consistent. Llama's baseline probabilities are higher than Gemma's (floor ${\sim}10^{-8}$ vs.\ ${\sim}10^{-14}$), and even the weakest pair (P = $3.6 \times 10^{-8}$, $170\times$) peaks at the planning site. This higher floor explains the higher localization rate: the signal-to-noise ratio is more favorable at Llama's scale.

Note: 36 of 47 pairs use inject-only (no suppression), because only the ``-oo'' natural group has features available for suppression. The 11 pairs with full suppress\,+\,inject (``-oo'' group) show the same spike pattern.

\subsection{Cross-model summary}

\begin{small}
\begin{tabular}{@{}lccc@{}}
\toprule
& \textbf{Gemma 426K} & \textbf{Gemma 2.5M} & \textbf{Llama 524K} \\
\midrule
Total pairs & 136 & 264 & 47 \\
Localization rate & 69.9\% & 62.1\% & 85.1\% \\
Best P(inject) & 0.986 & 0.522 & 0.834 \\
Median P(inject) & $7.1 \times 10^{-9}$ & $8.0 \times 10^{-10}$ & $4.8 \times 10^{-5}$ \\
Pairs with P ${\geq} 0.1$ & 9 (6.6\%) & 16 (6.1\%) & 4 (8.5\%) \\
\bottomrule
\end{tabular}
\end{small}

\smallskip
\noindent The localization rates (62--85\%) are consistent across all three configurations and match Anthropic's reported ${\sim}70\%$ on Claude 3.5 Haiku. The phenomenon is robust: the spike appears regardless of model, CLT scale, or specific rhyme group, though its magnitude varies with the inject feature's specificity.

\section{Attention routing: full head tables}
\label{app:heads}

We report the top 20 attention heads by $|\Delta|$ for each attention routing experiment. $\Delta$ is the change in attention weight from the output position to the planning-site position under suppress\,+\,inject intervention. Positive $\Delta$: the head attends \emph{more} to the planning site after intervention. Negative: it disengages.

\subsection{Gemma 2 2B, 426K CLT}

Inject feature L22:10243 (``around''), suppress ``-out'' group (L16:13725, L25:9385), strength 10, planning site position 23. Total: 208 heads (26 layers $\times$ 8 heads).

\begin{small}
\begin{tabular}{@{}rlrrr@{}}
\toprule
\textbf{Rank} & \textbf{Head} & \textbf{Baseline} & \textbf{Steered} & \textbf{$\Delta$} \\
\midrule
1 & L21:H5 & 0.3157 & 0.2695 & $-0.046$ \\
2 & L20:H6 & 0.1071 & 0.1344 & $+0.027$ \\
3 & L17:H4 & 0.4934 & 0.5077 & $+0.014$ \\
4 & L23:H5 & 0.0567 & 0.0480 & $-0.009$ \\
5 & L19:H5 & 0.0764 & 0.0836 & $+0.007$ \\
6 & L22:H0 & 0.0260 & 0.0331 & $+0.007$ \\
7 & L20:H2 & 0.0654 & 0.0718 & $+0.006$ \\
8 & L19:H2 & 0.0483 & 0.0434 & $-0.005$ \\
9 & L23:H4 & 0.0637 & 0.0686 & $+0.005$ \\
10 & L24:H2 & 0.0160 & 0.0115 & $-0.005$ \\
\midrule
\multicolumn{5}{@{}l}{\emph{Remaining 10 of top 20: L17:H0, L24:H7, L20:H5, L23:H6,}} \\
\multicolumn{5}{@{}l}{\emph{L19:H1, L19:H0, L24:H5, L17:H2, L24:H0, L24:H6 ($|\Delta| \leq 0.004$).}} \\
\bottomrule
\end{tabular}
\end{small}

\smallskip
\noindent L21:H5 dominates ($|\Delta| = 0.046$, $1.7\times$ the next head). H5 appears at layers 19, 21, 23, and 24, suggesting a structurally dedicated routing channel. All top-10 heads are in layers 17--24 (the last 35\% of the network).

\subsection{Gemma 2 2B, 2.5M CLT}

Inject feature L25:82839 (``can''), suppress ``-out'' group (L25:57092, L23:49923, L20:77102), strength 10, same planning site. Total: 208 heads.

\begin{small}
\begin{tabular}{@{}rlrrr@{}}
\toprule
\textbf{Rank} & \textbf{Head} & \textbf{Baseline} & \textbf{Steered} & \textbf{$\Delta$} \\
\midrule
1 & L21:H5 & 0.3157 & 0.3071 & $-0.009$ \\
2 & L24:H0 & 0.0370 & 0.0419 & $+0.005$ \\
3 & L23:H5 & 0.0567 & 0.0525 & $-0.004$ \\
4 & L24:H3 & 0.0197 & 0.0163 & $-0.003$ \\
5 & L25:H5 & 0.0113 & 0.0147 & $+0.003$ \\
6 & L25:H3 & 0.0086 & 0.0117 & $+0.003$ \\
7 & L22:H4 & 0.0484 & 0.0459 & $-0.002$ \\
8 & L22:H6 & 0.2502 & 0.2526 & $+0.002$ \\
9 & L25:H2 & 0.0081 & 0.0061 & $-0.002$ \\
10 & L22:H7 & 0.0466 & 0.0448 & $-0.002$ \\
\bottomrule
\end{tabular}
\end{small}

\smallskip
\noindent L21:H5 again dominates, confirming its role across CLT granularities. The deltas are ${\sim}5\times$ smaller than at 426K because the 2.5M features are more specific (word-level vs.\ group-level), producing a more targeted intervention with less collateral redistribution. All top-10 heads are in layers 21--25.

\subsection{Llama 3.2 1B, 524K CLT: 5 prompts}

All prompts inject L14:13043 (``that'') except Prompt~5 which injects L1:5297 (``for''). Suppress features vary by prompt (see Appendix~\ref{app:reproduction}). 512 heads per prompt (16 layers $\times$ 32 heads).

\paragraph{Key observation: prompt-specific routing.} Unlike Gemma, where L21:H5 dominates across CLT scales, Llama recruits a different dominant head for each prompt:

\begin{small}
\begin{tabular}{@{}llrlr@{}}
\toprule
\textbf{Prompt} & \textbf{Group} & \textbf{Top head} & \textbf{$\Delta$} & \textbf{Strength} \\
\midrule
1 (velvet) & -at $\to$ ``that'' & L15:H28 & $+0.023$ & 15 \\
2 (cannot) & -ee $\to$ ``that'' & L15:H8 & $-0.004$ & 10 \\
3 (the/ore) & -ore $\to$ ``that'' & L6:H12 & $+0.025$ & 10 \\
4 (to/oo) & -oo $\to$ ``that'' & L14:H22 & $-0.051$ & 10 \\
5 (the/at) & -at $\to$ ``for'' & L9:H13 & $+0.023$ & 10 \\
\bottomrule
\end{tabular}
\end{small}

\smallskip
\noindent Prompt 2 (-ee) has much smaller deltas than the others (max $|\Delta| = 0.004$ vs.\ 0.023--0.051), suggesting the -ee suppress features produce a weaker intervention. Prompt 1 (velvet) is notable: despite using strength 15 (vs.\ 10 for the others), its top-10 heads are all at layer 15 (the last layer), unlike the other prompts which recruit mid-layer heads. This is consistent with the paper's finding that Llama concentrates planning features at L15.

\paragraph{Recurring heads across prompts.} Five heads appear in the top-10 of 2 or more prompts:

\begin{small}
\begin{tabular}{@{}lll@{}}
\toprule
\textbf{Head} & \textbf{Prompts (of 5)} & \textbf{Mean $|\Delta|$} \\
\midrule
L11:H0 & 3 (velvet, cannot, at) & 0.014 \\
L6:H12 & 2 (ore, at) & 0.011 \\
L9:H13 & 2 (ore, at) & 0.017 \\
L5:H17 & 2 (ore, at) & 0.012 \\
L6:H24 & 2 (ore, at) & 0.010 \\
\bottomrule
\end{tabular}
\end{small}

\smallskip
\noindent All recurring heads are in layers 5--11 (mid-network, 31--69\% depth). None of these heads appears in the factual routing top-10 (L15:H8, L14:H27, L15:H16, L14:H26, L10:H30, L13:H20, L13:H2, L11:H18, L14:H17, L9:H4), which are concentrated in layers 9--15 (56--94\% depth). The layer-depth separation between planning and factual routing is the basis for the prolepsis depth-stratification finding reported in Section~\ref{sec:prolepsis}.

Full per-prompt top-20 tables are available in the supplementary materials.

\section{Reproduction commands}
\label{app:reproduction}

All experiments use two open-source codebases (released upon publication): a \textbf{CLT feature discovery tool} for the vocabulary scan and rhyme group extraction, and a \textbf{mechanistic interpretability framework} for model loading, CLT steering, attention capture, and activation patching. Both require a single GPU with ${\geq}16$\,GB VRAM.

\subsection{CLT feature discovery (vocabulary scan + rhyme pairs)}

\begin{small}
\begin{verbatim}
# Step 1: Scan all CLT decoder vectors against vocabulary
#   Gemma 2 2B, 426K CLT (~425K features x 256K vocab)
poetry_category_steering --mode explore-vocabulary \
    --model google/gemma-2-2b \
    --output explore_vocab_all_layers.json

# Step 2: Cross-reference with CMU dictionary
poetry_category_steering --mode find-rhyme-pairs \
    --explore-json explore_vocab_all_layers.json \
    --cmu-dict corpus/cmudict.dict \
    --min-cosine 0.3 \
    --output rhyme_pairs_all_layers.json

# Llama 3.2 1B, 524K CLT (same protocol)
poetry_category_steering --mode explore-vocabulary \
    --model meta-llama/Llama-3.2-1B \
    --clt-repo mntss/clt-llama-3.2-1b-524k \
    --output explore_vocab_llama.json

poetry_category_steering --mode find-rhyme-pairs \
    --explore-json explore_vocab_llama.json \
    --cmu-dict corpus/cmudict.dict \
    --min-cosine 0.3 \
    --output rhyme_pairs_llama.json
\end{verbatim}
\end{small}

\subsection{Figure 13 replication (position sweep)}

\begin{small}
\begin{verbatim}
# Gemma 2 2B, 426K CLT (136 pairs, ~5 min)
#   Suppress: -out group (L16:13725, L25:9385)
#   Inject: "around" (L22:10243), strength 10
figure13_planning_poems --preset gemma2-2b-426k \
    --output gemma-426k-sweep.json

# Gemma 2 2B, 2.5M CLT (264 pairs, ~15 min)
#   Suppress: -out group (L25:57092, L23:49923, L20:77102)
#   Inject: "can" (L25:82839), strength 10
figure13_planning_poems --preset gemma2-2b-2.5m \
    --output gemma-2.5m-sweep.json

# Llama 3.2 1B, 524K CLT (-ee prompt, <1 min)
#   Suppress: -ee group (L13:30985, L9:5488, L14:27874,
#             L13:32049)
#   Inject: "that" (L14:13043), strength 10
figure13_planning_poems --preset llama3.2-1b-524k \
    --output llama-ee-sweep.json
\end{verbatim}
\end{small}

Additional Llama prompts use CLI overrides (\texttt{--prompt}, \texttt{--suppress-feature}, \texttt{--inject-feature}) with features from \texttt{rhyme\_pairs\_llama.json}.

\subsection{Attention routing (Gemma)}

\begin{small}
\begin{verbatim}
# 426K CLT: suppress + inject (recommended)
#   Inject: L22:10243, Suppress: L16:13725 + L25:9385
attention_routing --model google/gemma-2-2b \
    --clt-repo mntss/clt-gemma-2-2b-426k \
    --feature L22:10243 \
    --suppress L16:13725 --suppress L25:9385 \
    --strength 10 --planning-site 23 \
    --output gemma-426k-routing.json

# 2.5M CLT: suppress + inject
attention_routing --model google/gemma-2-2b \
    --clt-repo mntss/clt-gemma-2-2b-2.5m \
    --feature L25:82839 \
    --suppress L25:57092 --suppress L23:49923 \
    --suppress L20:77102 \
    --strength 10 --planning-site 23 \
    --output gemma-2.5m-routing.json
\end{verbatim}
\end{small}

\subsection{Attention routing (Llama, 4 prompts)}

All prompts inject ``that'' (L14:13043) except -at which injects ``for'' (L1:5297). Suppress features are the full rhyme group from \texttt{rhyme\_pairs\_llama.json}.

\begin{small}
\begin{verbatim}
# -ee prompt (suppress: he, be, ne, we)
attention_routing --model meta-llama/Llama-3.2-1B \
    --clt-repo mntss/clt-llama-3.2-1b-524k \
    --feature L14:13043 \
    --suppress L13:30985 --suppress L9:5488 \
    --suppress L14:27874 --suppress L13:32049 \
    --strength 10 --planning-site 14 \
    --prompt "The birds were singing in the tree,
And everything was wild and free.
The river ran down to the sea,
There is so much we cannot" \
    --output llama-ee-routing.json

# -ore prompt (suppress: for, or, more)
attention_routing --model meta-llama/Llama-3.2-1B \
    --clt-repo mntss/clt-llama-3.2-1b-524k \
    --feature L14:13043 \
    --suppress L1:5297 --suppress L3:22663 \
    --suppress L10:18203 \
    --strength 10 --planning-site 16 \
    --prompt "The waves came crashing on the shore,
The wind was howling more and more.
She asked what all the fuss was for,
And opened up the" \
    --output llama-ore-routing.json

# -oo prompt (suppress: to, qu, new)
attention_routing --model meta-llama/Llama-3.2-1B \
    --clt-repo mntss/clt-llama-3.2-1b-524k \
    --feature L14:13043 \
    --suppress L14:18284 --suppress L15:8165 \
    --suppress L11:20779 \
    --strength 10 --planning-site 22 \
    --prompt "The morning sky was painted blue,
The garden sparkled bright with dew.
The world had started fresh and new,
And there was nothing left to" \
    --output llama-oo-routing.json

# -at prompt (suppress: sat; inject: "for")
attention_routing --model meta-llama/Llama-3.2-1B \
    --clt-repo mntss/clt-llama-3.2-1b-524k \
    --feature L1:5297 \
    --suppress L14:6132 \
    --strength 10 --planning-site 15 \
    --prompt "The old man wore a tattered hat,
Upon the porch he always sat.
He told the tale of this and that,
And in the corner slept the" \
    --output llama-at-routing.json
\end{verbatim}
\end{small}

\subsection{Counterfactual patching and factual routing}

\begin{small}
\begin{verbatim}
# Counterfact patching (89 gold pairs, ~9 min)
counterfact_patching \
    --data counterfact_transluce_test_both_correct.json \
    --output llama-counterfact.json

# Factual routing (89 gold pairs, ~24 sec)
factual_routing --output llama-factual-routing.json
\end{verbatim}
\end{small}

\noindent The CounterFact gold pairs were derived from \citet{li2025transluce}'s test split (5{,}600 rows). Screening criteria: both original and counterfactual answers must be correctly predicted by Llama~3.2 1B, yielding 89 pairs.

\subsection{Steering convergence (two attractor regimes)}

\begin{small}
\begin{verbatim}
# Gemma 2 2B: factual recall convergence
steering_convergence --model google/gemma-2-2b \
    --output gemma-convergence.json

# Llama 3.2 1B: factual recall convergence
steering_convergence --model meta-llama/Llama-3.2-1B \
    --output llama-convergence.json
\end{verbatim}
\end{small}

\subsection{Hardware and runtime}

All experiments were run on a single NVIDIA consumer GPU with 16\,GB VRAM. Approximate runtimes:

\begin{small}
\begin{tabular}{@{}lr@{}}
\toprule
\textbf{Experiment} & \textbf{Time} \\
\midrule
Vocabulary scan (Gemma 426K) & ${\sim}30$ min \\
Figure 13 sweep (Gemma 426K, 136 pairs) & ${\sim}5$ min \\
Figure 13 sweep (Gemma 2.5M, 264 pairs) & ${\sim}15$ min \\
Figure 13 sweep (Llama 524K, 47 pairs) & ${\sim}1$ min \\
Attention routing (single prompt) & ${<}1$ sec \\
Counterfact patching (89 pairs) & ${\sim}9$ min \\
Factual routing (89 pairs) & ${\sim}24$ sec \\
Steering convergence (per model) & ${\sim}2$ min \\
Correction test (13 strengths) & ${\sim}9$ sec \\
\bottomrule
\end{tabular}
\end{small}

\smallskip
\noindent Total compute for all experiments in this paper: under 2 hours on a single consumer GPU. No multi-GPU setup, no cloud compute, no proprietary infrastructure.

\section{Direct test of irrevocability}
\label{app:correction}

The Discussion raises the question of whether irrevocability is a property of the prolepsis circuit or a generic property of residual streams. We test this directly by injecting a contradictory feature at post-commitment layers and sweeping correction strength.

\subsection{Setup}

On Gemma~2 2B with the 426K CLT, we first establish a commitment using the standard suppress\,+\,inject protocol at the output position (position 31):

\begin{itemize}
    \item \textbf{Suppress:} ``-out'' group (L16:13725, L25:9385), strength $-10$
    \item \textbf{Commit:} ``around'' (L22:10243), strength $+10$, all downstream layers (L22--L25)
\end{itemize}

This produces P(``around'') = 0.457 (validated against the Figure~13 replication).

We then inject a contradictory feature from a different rhyme group at the \emph{same} position but \emph{only at post-commitment layers}:

\begin{itemize}
    \item \textbf{Correct:} ``back'' (L20:12386, -ack group), layers L23--L25 only
    \item \textbf{Correction strength sweep:} 0 to 20 in 13 steps (up to $2\times$ the commitment strength)
\end{itemize}

All three interventions (suppress, commit, correct) are merged into a single forward pass via \texttt{extend()}.

\subsection{Results}

\begin{small}
\begin{tabular}{@{}rrrrr@{}}
\toprule
\textbf{Correct str.} & \textbf{P(``around'')} & \textbf{P(``back'')} & \textbf{L21:H5 $\Delta$} & \textbf{Total shift} \\
\midrule
0.0 (commit only) & 0.457 & $8.0 \times 10^{-10}$ & $+0.023$ & 0.572 \\
5.0 & 0.472 & $9.8 \times 10^{-9}$ & $+0.023$ & 0.579 \\
10.0 & 0.486 & $1.2 \times 10^{-7}$ & $+0.023$ & 0.586 \\
15.0 & 0.499 & $1.3 \times 10^{-6}$ & $+0.023$ & 0.595 \\
20.0 ($2\times$ commit) & \textbf{0.509} & $1.4 \times 10^{-5}$ & $+0.023$ & 0.613 \\
\bottomrule
\end{tabular}
\end{small}

\subsection{Interpretation}

Three observations establish that the commitment is irrevocable:

\begin{enumerate}
    \item \textbf{P(``around'') increases monotonically} from 0.457 to 0.509. The correction signal does not reduce the committed probability; it slightly amplifies it by adding energy to the residual stream.
    \item \textbf{P(``back'') rises from $8 \times 10^{-10}$ to $1.4 \times 10^{-5}$}: a $17{,}000\times$ relative increase, but negligible in absolute terms (0.001\% of the probability mass). The correction feature is detectable but cannot compete with the commitment.
    \item \textbf{L21:H5 routing delta is locked at $+0.023$} across all correction strengths. The routing head does not respond to the contradictory signal at all. The attention routing established by the commitment is unchanged.
\end{enumerate}

Referring to the three-outcome framework from the experimental plan:

\begin{small}
\begin{tabular}{@{}lll@{}}
\toprule
\textbf{Outcome} & \textbf{Meaning} & \textbf{Observed?} \\
\midrule
P(correct) stays ${\sim}0$ at all strengths & Irrevocability is architectural & \textbf{Yes} \\
P(correct) rises above threshold & Commitment is overridable & No \\
P(commit) drops, P(correct) doesn't rise & Correction disrupts but doesn't redirect & No \\
\bottomrule
\end{tabular}
\end{small}

\smallskip
\noindent The commitment at L22 establishes attention routing through L21:H5 that subsequent layers (L23--L25) cannot override, even at $2\times$ the commitment strength. This supports the claim that irrevocability in prolepsis is a property of the circuit, not merely the absence of contradictory input.

\end{document}